\documentclass{article}

\usepackage[final]{neurips_2024}

\usepackage[utf8]{inputenc} % allow utf-8 input
\usepackage[T1]{fontenc}    % use 8-bit T1 fonts
\usepackage{hyperref}       % hyperlinks
\usepackage{url}            % simple URL typesetting
\usepackage{booktabs}       % professional-quality tables
\usepackage{amsfonts}       % blackboard math symbols
\usepackage{nicefrac}       % compact symbols for 1/2, etc.
\usepackage{microtype}      % microtypography
\usepackage{xcolor}         % colors
\usepackage{graphicx} % Add this line to import the necessary package for including graphics

\usepackage{amsmath}
\usepackage{subfigure}
\usepackage{caption}

\usepackage{wrapfig}
\usepackage[toc,page,title]{appendix}
\usepackage{minitoc}
\usepackage[noend]{algpseudocode}
\usepackage{algorithm}
\usepackage{booktabs}

\newcommand{\tool}{MUN}
\newcommand{\toolhp}{MUN-noDAD}

\title{Learning World Models for Unconstrained Goal Navigation}

\author{%
  Yuanlin Duan\\
  Rutgers University\\
  \texttt{yuanlin.duan@rutgers.edu} \\
  \And
  Wensen Mao \\
  Rutgers University \\
  \texttt{wm300@cs.rutgers.edu} \\
  \And
  He Zhu \\
  Rutgers University \\
  \texttt{hz375@cs.rutgers.edu} \\
}

\begin{document}
\include{pythonlisting}

\maketitle

\begin{abstract}
  Learning world models offers a promising avenue for goal-conditioned reinforcement learning with sparse rewards. By allowing agents to plan actions or exploratory goals without direct interaction with the environment, world models enhance exploration efficiency. The quality of a world model hinges on the richness of data stored in the agent's replay buffer, with expectations of reasonable generalization across the state space surrounding recorded trajectories. However, challenges arise in generalizing learned world models to state transitions backward along recorded trajectories or between states across different trajectories, hindering their ability to accurately model real-world dynamics. To address these challenges, we introduce a novel goal-directed exploration algorithm, \tool{} (short for "World Models for Unconstrained Goal Navigation"). This algorithm is capable of modeling state transitions between arbitrary subgoal states in the replay buffer, thereby facilitating the learning of policies to navigate between any "key" states. Experimental results demonstrate that \tool{} strengthens the reliability of world models and significantly improves the policy's capacity to generalize across new goal settings.
\end{abstract}

\section{Introduction}

\begin{wrapfigure}[12]{r}{0.6\textwidth}
\vspace{-0cm}
  \centering
  \includegraphics[width=0.6\textwidth]{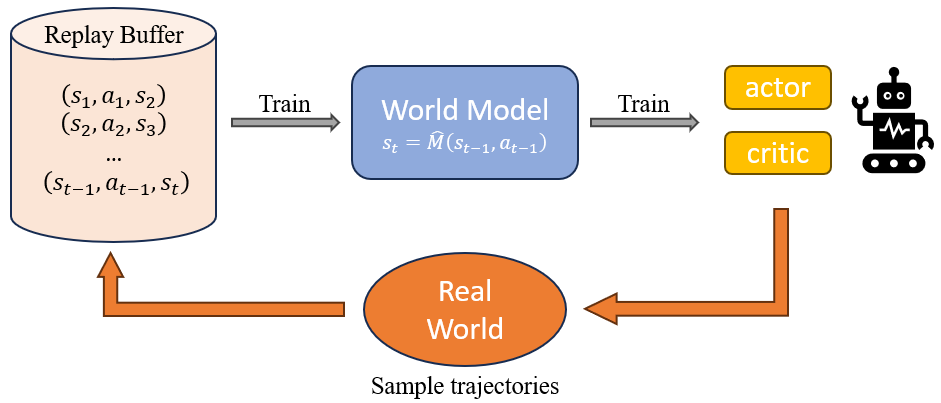}
  \caption{The general framework of model-based RL.}
  \label{fig:general_framework_of_mbrl}
\end{wrapfigure}

Goal-conditioned reinforcement learning (GCRL) has emerged as a powerful framework for learning diverse skills within an environment and subsequently solving tasks based on user-specified goal commands, without requiring further training(~\cite{mendonca2021discovering, andrychowicz2017hindsight}). Given that specifying dense task rewards for GCRL requires domain expertise, access to object positions, is time-consuming, and is prone to human errors, rewards in GCRL are typically sparse, signaling success only upon reaching goal states. However, sparse rewards pose a challenge for exploration during training. To address this challenge, several previous methods, e.g., ~\cite{hafner2019dream,hansen2023td, mendonca2021discovering} have proposed learning a generative world model of the environment using a reconstruction (decoder) objective, an instantiation of Model-based Reinforcement Learning (MBRL), visualized in Fig.~\ref{fig:general_framework_of_mbrl}. This approach is appealing because the world model can provide a rich learning signal(~\cite{yu2020mopo, georgiev2024pwm}). For example, world models allow agents to plan their actions or exploratory goals without directly interacting with the real environment for more efficient exploration(~\cite{hu2023planning,sekar2020planning}). %World models particularly enable agents to solve many different tasks by learning from imagined experiences, leading to better generalization to new situations.

%The key challenge in improving the performance of GCRL is exploration - how can GCRL agents discovering new goals and learning to reliably achieve them? identifying goal commands capable of generating novel states with high expected uncertainty or information gain through imagined rollouts of the world model.

Existing MBRL techniques train world models to capture the dynamics of the environment from the agent's past experiences stored in a replay buffer. The richness of the data stored in the agent's replay buffer directly impacts the quality of a World Model. It is expected that the world model generalizes reasonably well to the state space surrounding the trajectories recorded in the replay buffer. However, the world model may not generalize well to state transitions backward along recorded trajectories or to states across different trajectories, which impedes the world model's learning of the real-world dynamics. 

%Through an analysis of the internal structure of current Model-based RL replay buffers, it has been observed that the dynamic transitions therein are often unidirectional, 

To induce a data-rich replay buffer covering a wide range of dynamic transitions,
in this paper, we present a novel goal-directed exploration algorithm for effective world modeling and policy learning, \tool{} (short for "World \textbf{M}odels for \textbf{U}nconstrained Goal \textbf{N}avigation").
%\tool{} supports modeling the state transition between arbitrary subgoal states in the replay buffer no matter backward from recored trajectoies or jump to states on different trajectores 
\tool{} facilitates modeling state transitions between any subgoal states in the replay buffer, whether tracing back along recorded trajectories or transitioning between states on separate trajectories. This enhances the reliability of the learned world model and significantly improves the generalizability of the policy derived from the model to real-world environments, thereby boosting the exploration capabilities of the method. Additionally, we introduce a simple and practical strategy for discovering \emph{key} subgoal states from the replay buffer. The key subgoals precisely mark the milestones necessary for task completion, such as steps like grasping and releasing blocks in the context of block-stacking scenarios. By world modeling and policy learning for unconstrained navigation between these key states, \tool{} can generalize to new goal settings, such as block unstacking that was not given to the agent at training time.

%, known as APS (action plan subgoals).

%which is utilized to construct the Two-direction Replay Buffer. These two techniques can also be widely applied in Model-free RL methods.

%Doing so significantly enhances the richness of dynamic transitions in the replay buffer, thereby accelerating the comprehensive learning of the real environment by the World Model in terms of both depth and breadth. 

Our key contributions are as follows. First, we propose a novel goal-directed exploration algorithm \tool{} for effective world modeling of state transition between arbitrary subgoal states in replay buffers. As the quality of the world model improves, \tool{} becomes highly effective at learning goal-conditioned policies that excel at exploration in sparse-reward environments. Second, we present a practical strategy for identifying pivotal subgoal states, which serve as milestones in completing sophisticated tasks. By training world models for unconstrained transition between these milestones, our method enables learning policies that can adapt to novel goal scenarios. Finally, we evaluate \tool{} in challenging robotics environments, such as guiding a multi-legged ant robot through a maze, maneuvering a robot arm amidst cluttered tabletop objects, and rotating items in the grasp of an anthropomorphic robotic hand. Across these environments, \tool{} exhibits superior efficiency in training generalizable goal-conditioned policies compared to baseline methods and ablations.

\section{Problem Setup and Background}
\label{sec:psb}

We consider the problem of goal-conditioned reinforcement learning (GCRL) under a Markov Decision Process (MDP) parameterized by \((S, A, P, G, \eta, R, \rho_0)\). \(S\) and \(A\) are the state and action spaces, respectively. The probability distribution of the initial states is given by \(\rho_0(s)\), and \(P(s'|s, a)\) is the transition probability. \(\eta : S \to G\) is a mapping from the state space to the goal space, which assumes that every state \(s\) can be mapped to a corresponding achieved goal \(g\). The reward function \(R\) is defined as \(R(s, a, s', g) = 1\{\eta(s') = g\}\). We assume that each episode has a fixed horizon \(T\). For ease of presentation, we further assume $S = G$ and $\eta$ is an identify function in this paper.

A goal-conditioned policy is a probability distribution \(\pi : S \times G \times A \to \mathbb{R}^+\), which gives rise to trajectory samples of the form \(\tau = \{s_0, a_0, g, s_1, \ldots, s_T\}\). The purpose of the policy \(\pi\) is to learn how to reach the goals drawn from the goal distribution \(p_g\). With a discount factor \(\gamma \in (0, 1)\), it maximizes \(J(\pi) = \mathbb{E}_{g \sim p_g, \tau \sim \pi(g)} \left[ \sum_{t=0}^{T-1} \gamma^t \cdot R(s_t, a_t, s_{t+1}, g) \right]\).

In the context of model-based reinforcement learning (MBRL), a world model $\hat{M}$ is trained over trajectories sampled from the agent's interactions with the real environment, which are stored in a replay buffer, to predict the dynamics of the real environment. Fig.~\ref{fig:general_framework_of_mbrl} illustrates the general MBRL framework. 
%Learning a world model involves collecting trajectories sampled from the policy in the real environment and storing them in a replay buffer. 
%The replay buffer is then used to train a World Model 
We use the world model structure $\hat{M}$ of Dreamer (\cite{hafner2019dream, hafner2019learning, hafner2020mastering, hafner2023mastering}) to learn real environment dynamics as a recurrent state-space model (RSSM). We provide a detailed explanation of the network architecture and working principles of the RSSM in Appendix~\ref{subs: rssm}. Our study focuses on tackling the world model learning problem in goal-conditioned model based reinforcement learning settings. 
%We focus on the goal-conditioned RL tasks, where the agent is trained to achieve a set of predefined goals in the environment.
Particularly, we consider \textbf{GC-Dreamer} (goal-conditioned Dreamer) as an important baseline with the following learning components:
\begin{equation}\label{eq:actor_critic}
  \text{World Model:\quad} \hat{M}(s_t | s_{t-1}, a_{t-1}) \qquad \text{Actor:\quad} \pi^G(a_t | s_t, g) \qquad \text{Critic:\quad} V(s_t, g)
\end{equation}
In GC-Dreamer, the goal-conditioned agent $\pi^G(a | s, g)$ samples goal commands $g \in G$ from the given environment goal distribution $p_g$ to collect trajectories in the real world. These trajectories are used to train the world model $\hat{M}$, and subsequently, $\pi^G$ is trained on imagined rollouts generated by $\hat{M}$ using the model-based actor-critic algorithm in Dreamer~\cite{hafner2020mastering}, with these two steps run in alternation. The critic estimates the sum of future rewards $\sum_t r^G_t$, and the actor tries to maximize the predicted values from the critic. The goal-reaching reward $r^G$ is defined by the self-supervised temporal distance network $D_t$ (\cite{mendonca2021discovering}), i.e. $r^G(s, g) = -D_t(s, g)$. $D_t$ predicts the anticipated number of action steps needed to transition from $s$ to $g$. Essentially, $\pi^G$ is reinforced to minimize the action steps required to transition from the current state $s$ to a sampled goal state $g$. 
The temporal distance estimator $D_t$ is trained by extracting pairs of states $s_t$ and $s_{t+k}$ from an imagined rollout generated by running the policy over the world model and predicting the distance $k$ between them as follows:
\begin{equation}\label{eq: temporal_distance}
  D_t\big(\Psi(s_{t}), \Psi(s_{t+k})\big) \approx k/H
\end{equation}
Here, $\Psi$ represents the preprocessing for imagined states, such as transforming them into the world model's latent space (we assume $S = G$ in the paper). $H$ represents the total length of the imagined rollout. Further details on the training procedure of $D_t$ can be found in Appendix~\ref{subs: Dt-training}.

\section{Training World Models for Unconstrained Goal Navigation}

In this section, we introduce \tool{}, our main approach to addressing the core challenge in GCRL: efficient exploration in long-horizon, sparse-reward environments. Our approach focuses on enhancing the agent's understanding of the real-world environment through improved dynamic (world) modeling and latent space representation. As the quality of the world model improves, the goal-conditioned policy developed from it generalizes more effectively to the real environment. By closing the generalization gap between the policy's behavior in the real environment and the world model, \tool{} effectively guides the agent's exploration towards the desired goal region in the real environment.

\subsection{Training Generalizable World Models}

\begin{figure}[t]
    \centering
    \subfigure[Key subgoal states in a 3-Block Stacking task.]{\includegraphics[width=0.46\textwidth]{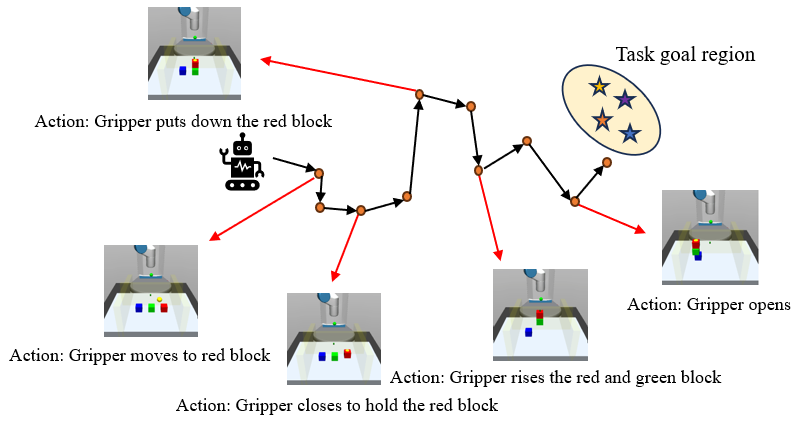}\label{fig:keystate}}
    \hspace{0.5cm}
    \subfigure[Bidirectional Replay Buffer]{\includegraphics[width=0.48\textwidth]{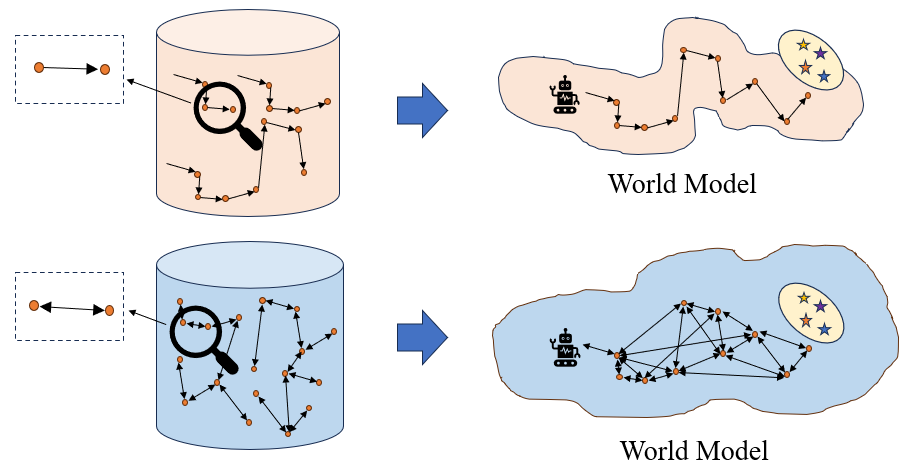}\label{fig:two-direction-RB}}
  \caption{In Fig.~\ref{fig:keystate}, we illustrate the key states involved in completing the task of 3-block stacking. 
  In Fig.~\ref{fig:two-direction-RB}, we demonstrate the significant advantages of the bidirectional replay buffer used in \tool{} over traditional methods in learning world models. }
\end{figure}

%In model-based RL, the effectiveness of policy learning is highly dependent on how well the world model represents the real world. Therefore, our key idea is to train a world model that accurately characterizes the real-world environment.
%As in Model-based RL the learning of a policy heavily relies on the degree to which the World Model learns about the real world, our key idea is to train a world model that accurately characterize the real-world environment. 
Fig~\ref{fig:general_framework_of_mbrl} illustrates the general framework of Model-based RL, where world models are trained using agent's experiences stored in a replay buffer populated with observed environment transitions $(s_t, a_t, s_{t+1})$ linking the environment's future states $s_{t+1}$ and past states $s_t$ along with the corresponding control actions $a_t$. The richness of the environment space and dynamic transitions captured by the replay buffer define the extent of what a world model can learn about the real environment. Through supervised learning, the model can generalize reasonably well within the state space moving \emph{forward} along the trajectories recorded in the replay buffer. However, it may be inaccurate for the state transitions moving \emph{backward} along the recorded trajectories or \emph{across} different trajectories. Consider the task of stacking blocks using a robot manipulator in Fig.~\ref{fig:keystate}. When humans learn to stack blocks, they also understand how to reverse the process to unstack the blocks or return to the initial state. In contrast, a world model trained solely on data from policies under training for stacking is unlikely to accurately model the unstacking process. As a result, the model may yield hallucinated trajectories for training policies, causing a significant discrepancy between the policy's behavior in the model and in the real world, thereby leading to ineffective exploration.

To improve model generalizability, in \tool{}, we proposed to learn world models capable of characterizing state transitions between any states in the replay buffer, whether by tracing back along recorded trajectories or transitioning between states on separate trajectories. %As the generalization gap of a model to the real-world environment reduces, \tool{} can effectively leverage the world model for effective exploration towards the task goal region.    
%As such, the policy learned over the model (using richer learning signals e.g. Equation~\ref{eq: temporal_distance_reward4}) generalizes to the real-world environments, thus enhancing the exploration capability. 
Fig.~\ref{fig:two-direction-RB} visualizes the comparison between the bidirectional replay buffer for learning world models used in \tool{} and the unidirectional replay buffer in conventional model-based algorithms. The bidirectional replay buffer not only covers a wider observation space but also captures a richer set of dynamic transitions. As discussed in Sec.~\ref{sec:psb}, due to joint optimization, the richer set of dynamic transitions in \tool{} allows for a more reliable latent representation of the environmental space and consequently a higher quality reward function (Equation~\ref{eq: temporal_distance}) for training policies generalizable to the real environment on top of the learned model.
%can directly guide the optimization of the latent space. Consequently, the World Model trained with the bidirectional replay buffer achieves 

%The bidirectional replay buffer can help the World Model not only learn how to simulate dynamic transitions from the initial state area to the target state area but also learn how to simulate dynamic transitions between arbitrary combinations of key state areas on the path. 

\begin{algorithm}[t]
\caption{The main training framework of \tool{}}
\label{Algorithm1-TrainingFrame}
\begin{algorithmic}[1]
    \State \textbf{Input:} Policy $\pi^G$, World Model $\hat{M}$, reward function $r^G$, subgoals transfer number $N_{s}$, subgoal time limit $T_{s}$
    \State Initialize buffers $D, D_{DAD}, D_{egc}$
    \For{$i = 1$ to $N_{train}$}
        \If{Should Plan Subgoals}
            \State $B_{egc} \leftarrow$ A batch of episodes from $D_{egc}$
            \State $G_{subgoals} \leftarrow$ DAD($B_{egc}$) with Algorithm \ref{Algorithm2-APS}
        \EndIf\label{alg:updatesubgoals}
        \State Initialize empty trajectory $\tau$
        \For{$s = 1$ to $N_{s}$}\label{alg:muntraverse}
            \State $t_s = 0$
            \State $g_s =$ Sample a subgoal randomly from $G_{subgoals}$
            \While{agent has not reached $g_s$ and $t_s < T_{s}$}
                \State Append one step in real environment with $\pi^G$ using goal $g_s$ to $\tau$
                \State $t_s \leftarrow t_s + 1$
            \EndWhile
        \EndFor
        \State $D_{DAD} \leftarrow D_{DAD} \cup \{\tau\}$
        \State $\tau'\ \gets$ Trajectory of $\pi^G$ sampled using the environment goal distribution $g \sim p_g$
        \State $D_{egc} \leftarrow D_{egc} \cup \tau'$ 
        \State $D \leftarrow D_{DAD} \cup D_{egc}$
        \State Update $\hat{M}$ with $D$ \label{alg:learnM}
        \State Update $\pi^G$ in imagination with $\hat{M}$ to maximize $r^G$ \label{alg:learnG}
    \EndFor
\end{algorithmic}
\end{algorithm}

We depict the learning algorithm in \tool{} in Algorithm~\ref{Algorithm1-TrainingFrame}. In the algorithm, we maintain $G_{subgoals}$ as a set of pivot subgoal states sampled from the relay buffer (illustrated in Algorithm~\ref{Algorithm2-APS}) and aim to learn world models capable of seamless transitions between these subgoals. At line~\ref{alg:updatesubgoals}, we periodically update $G_{subgoals}$ as the training evolves. 
%In the first stage, we randomly select a subgoal state from the set of key subgoal states, use a function $\eta$ to convert it to the goal space, 
In the loop starting from line~\ref{alg:muntraverse}, we repeatedly sample \( N_s \) subgoals from \( G_{\text{subgoals}} \) and direct the agent to sequentially reach these subgoals within a time limit of \( T_s \) steps for each. In this way, \tool{} samples a replay buffer that records bidirectional state transitions between the subgoals in $G_{subgoals}$. Based on our experience, we find that setting \( N_s = 2 \) is sufficient. Further discussion on the setting of \( N_s \) is provided in Sec~\ref{sec:experiments}. 
%Through a large number of such randomly sampled bidirectional trajectories of arbitrary subgoal combinations, we obtain the Bidirectional Replay Buffer. 
At line~\ref{alg:learnM}, we train the world model $\hat{M}$ using trajectories collected by both the goal commands from $G_{subgoals}$ (stored in $D_{DAD}$) and that sampled from the environment goal distribution $p_g$ (stored in $D_{egc}$). Then, we sample imaginary rollouts from the world model for policy training at line~\ref{alg:learnG}. 
%In our experience, we find it suffices to set $N_s = 2$. We will discuss more about the setting of $N_s$ in Sec~\ref{sec:experiments}.

%obtaining $g_{s1}$, and instruct the agent to reach it  
%In the second stage, using the same process, we select $g_{s2}$, allowing the agent to move from the endpoint of the first stage to $g_{s2}$. 
%$(g_{s1}, g_{s2})$ 
%can be any combination of key %subgoal states, and the number of subgoals within a combination can be expanded to three or more to obtain richer bidirectional trajectories. 

%Algorithm~\ref{Algorithm1-TrainingFrame} illustrates our method framework and Algorithm~\ref{Algorithm2-APS} shows how to find key subgoal states using the APS method.
 
%In this dependency, the prerequisite for the policy to learn to reach these key subgoal states is that the World Model must effectively encode the space surrounding these key subgoal states and accurately predict the dynamic transition functions between them.  Therefore, enhancing the efficiency and proficiency of World Model learning regarding the encoding of the space around key subgoal states and dynamic transition prediction  becomes one of the crucial challenges in Model-based RL.

%The replay buffer is populated by the agent using the policy to sample from the real environment. 
%In conventional Model-based approaches, these trajectories are often sampled with the environment's goals as targets, 

\textbf{Comparison with Go-Explore.} We highlight the key difference between \tool{}'s exploration strategy and the recently popular "Go-Explore" strategy(~\cite{ecoffet2019go,pislar2021should,tuyls2022multi,hu2023planning}), designed for exploration-extensive long-term GCRL settings. In Go-Explore, each training episode comprises two phases: the "Go-phase" and the "Explore-phase". During the "Go-phase," the goal-conditioned policy $\pi_G$ directs the agent to an "interesting" goal(~\cite{pong2019skew,pitis2020maximum} )(e.g., states with low frequency of occurrence in the replay buffer), resulting in a final state $s_{T_g}$ after $T_g$ steps. Following this, the "Explore-phase" begins, where an undirected exploration policy takes over from $s_{T_g}$ for the remaining $T_e$ timesteps. This exploration policy is trained to maximize an intrinsic exploration reward(~\cite{bellemare2016unifying,pathak2017curiosity,burda2018exploration,sekar2020planning}) (e.g., to visit areas of the real world that the World Model has not yet learned well). This structure of training episodes has been shown to result in richer exploration(~\cite{pislar2021should}). In \tool{}, when $N_s = 2$, Algorithm~\ref{Algorithm1-TrainingFrame} essentially replaces the "Explore-phase" in "Go-Explore" with another "Go-phase". Thus, the algorithm directs the agent to navigate between two "interesting" goals selected from the replay buffer. Firstly, \tool{} is computationally efficient as it eliminates the need to train a separate exploration policy and an ensemble of world models used to generate intrinsic exploration rewards. Secondly, \tool{} trains the world model for unconstrained navigation between goal states in the replay buffer, thereby improving the model's generalization to the real-world environment and leveraging the model for exploration. We empirically compare the two strategies in the context of model-based GCRL in Sec.~\ref{sec:experiments}.

\subsection{Key Subgoal Generation through Distinct Action Discovery (DAD)}

Having set up the main learning algorithm, we seek to address: how should we pick
exploration-inducing goals from the replay buffer at training time to help learn generalizable world models? A straightforward strategy is to sample trajectories from the replay buffer and select subgoal states at fixed time intervals along these trajectories. We improve this simple approach with a practical method called DAD (Distinct Action Discovery) for identifying \emph{key} subgoal states, which represent the pivotal milestones necessary to complete a complex task. Consider the block stacking task as an example. The robotic arm must be able to move its gripper to the vicinity of a block, close the gripper, lift the block, move to the top of another block, release the block, and finally open the gripper. These key subgoal states are essential for completing the task. We illustrate the roles of key states in a 3-Block Stacking task in Fig.~\ref{fig:keystate}. For an agent to learn this task, it must master reaching and navigating between the key states. By training world models for unconstrained transitions between these key states, \tool{} can develop models that more accurately capture the task structure and learn policies capable of adapting to novel goal scenarios e.g. block unstacking.

%In any given task, there inevitably exist certain states that are necessary for successfully completing the task. 

\begin{wrapfigure}[8]{r}{0.56\textwidth}
\vspace{-0.7cm}
\scalebox{0.8}{
\begin{minipage}{0.68\textwidth}
\begin{algorithm}[H]
\caption{Key Subgoal Generation by Distinct Action Discovery}
\label{Algorithm2-APS}
\begin{algorithmic}[1]
\State \textbf{Function:} DAD(...)
\State \textbf{Input:} A batch of episodes $B_{egc}$, number of subgoals $N_{subgoals}$
\State $A \leftarrow$ Find the most different $N_{subgoals}$ actions from $B_{egc}$ by FPS
\State $S_{subgoals} \leftarrow$ get the corresponding states of $A$ from $B_{egc}$
\State $G_{subgoals} \leftarrow$ $\eta(S_{subgoals})$
\State \textbf{return} $G_{subgoals}$
\end{algorithmic}
\end{algorithm}
\end{minipage}
}
\end{wrapfigure}

There exist methods for identifying key states(~\cite{zhang2021world,paul2019learning}). However, these methods often tend to be overly complex, leading to insufficient generalization across different environments and requiring adjustments to the methods' components or parameters for various tasks. Our approach is based on the observation that certain actions are crucial at different stages of task completion. For instance, in a block stacking task, the robotic arm must learn actions such as closing the gripper, grasping the block, lifting it, and releasing the gripper. When the agent performs these key actions, the corresponding states can often be considered key subgoal states. By selecting actions that significantly differ along trajectories and extracting the corresponding states during these actions, we can identify potential key subgoal states. The Farthest Point Sampling (FPS) algorithm(\cite{eldar1997farthest}) provides a simple and efficient method for selecting $N$ points with maximal differences from a set. We apply FPS(\cite{eldar1997farthest}) to choose $N$ time steps with the greatest variations in actions from a batch of trajectory data, thereby obtaining the set of key subgoal states corresponding to these time steps. Algorithm~\ref{Algorithm2-APS} shows how \tool{} finds key subgoal states using the DAD method.

\section{Experiments}
\label{sec:experiments}

We evaluate \tool{} across various robotic manipulation and navigation environments, aiming to address the following three research questions:
(RQ1) Does \tool{} outperform other goal-conditioned model-based reinforcement learning baselines with advanced exploration strategies?
(RQ2) Can DAD effectively identify key subgoal states along trajectories to the environment goal region?
(RQ3) Does \tool{} successfully leverage the bi-directional replay buffer to train a generalizable policy for navigating effectively between arbitrary subgoals?

\subsection{Environments}

\begin{figure}[t] 
  \centering
  \includegraphics[width=0.9\textwidth]{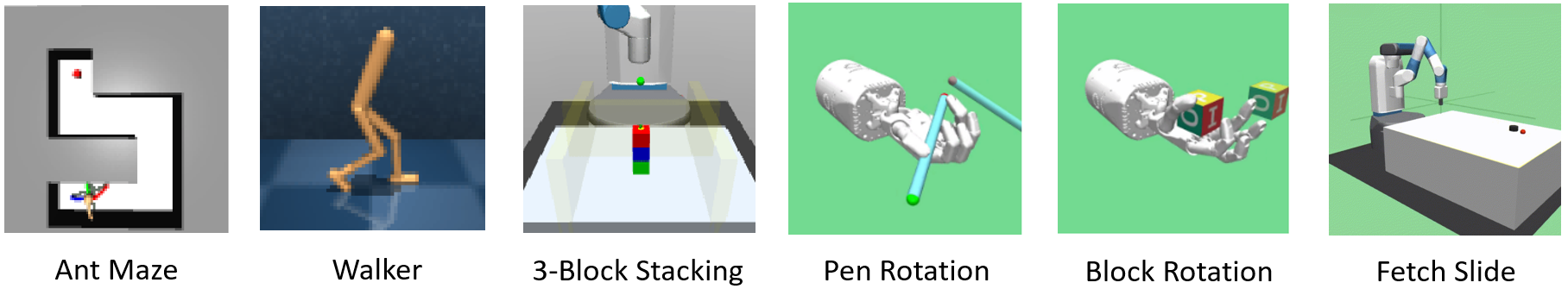}
  \caption{We evaluate \tool{} on 6 environments: Ant Maze, Walker, 3-Block Stacking, Block Rotation, Pen Rotation, Fetch Slide.}
  \label{fig:environments}
  \vspace{-15pt}
\end{figure}

%We use six challenging goal conditional tasks to do experiments on \tool{} and other baselines to answer the above questions. 

%In \textbf{3-Block Stacking}, a robotic arm equipped with a two-fingered gripper faces the challenge of navigating a tabletop environment scattered with three blocks of different colors and surrounded by boundary walls. This task is notorious for its complexity in robotic reinforcement learning, often requiring costly forms of supervision or extensive computational resources for effective exploration. Previous solutions have leaned on methods like demonstrations, curriculum learning, or massive amounts of simulator data, highlighting the formidable nature of the task. During evaluation, the agent must successfully stack the three blocks into a tower formation, varying both their positions and order, necessitating the mastery of pushing, picking, and stacking actions, as well as the discovery of intricate action paths to accomplish the task within the environment.
We conducted experiments on six challenging goal-conditioned tasks to evaluate \tool{}.
In \textbf{Ant-Maze}, an ant-like robot is tasked to learn complex 4-legged locomotion behavior and navigate around the hallways within a maze structure.
%challenged with mastering the intricacies of four-legged locomotion and navigating through narrow hallways within a maze structure. Following the PEG framework, the maze comprises three turns and four straight corridors, presenting a formidable navigation challenge.
The \textbf{Walker} task involves a two-legged robot learning to control its leg joints effectively to achieve stable walking to reach goals along a flat plane forward or backward.
In \textbf{3-Block Stacking}, a robot arm with a two-fingered gripper operates on a tabletop with three blocks. The goal is to stack the blocks into a tower configuration. The agent needs to learn to push, pick, and stack objects while discovering complex action sequences to complete the task in the environment. Previous solutions have relied on methods like demonstrations, curriculum learning, or extensive simulator data, highlighting the task's difficulty(~\cite{ecoffet2019go, li2020towards,nair2018overcoming, lanier2019curiosity}).
The \textbf{Block Rotation} and \textbf{Pen Rotation} tasks require the agent to manipulate a block and a pen, respectively, to achieve a randomly specified orientation along all axes. %Block Rotation focuses on rotating a block with random target rotations , while 
Pen Rotation is particularly challenging due to the pen's thinness, requiring precise control to prevent it from dropping. %Both tasks utilize variant versions within the gymnasium environment, featuring randomized target rotations for each episode to enhance learning variability.
In \textbf{Fetch Slide}, a manipulator slides a puck to a designated goal area on a slippery table.
%a robotic arm is equipped with a two-fingered gripper, and its task is to push a small object across a flat surface to reach a designated goal area. 
Unlike tasks that involve direct manipulation, Fetch Slide emphasizes the challenge of accurately controlling the force and direction of the push operation, as the puck must slide across the flat surface to the target. %The difficulty arises from the need for precise control. %, as the puck may not always respond predictably, particularly when accounting for factors like friction or object shape.
See Appendix.~\ref{sec: envs} for more information about environments.

\subsection{Baselines}

We compare \tool{} with the following baselines. The \textbf{GC-Dreamer} baseline is discussed in Sec.~\ref{sec:psb}.
%Dreamer is recently recognized as one of the strongest baselines in MBRL, we use the goal-conditioned dreamer 
%relized in the LEXA paper(\cite{mendonca2021discovering}) represented as \textbf{GC-Dreamer} to be one baseline.
%At each round sampling, the \textbf{GC-Dreamer} use the goal returned by environment as target of goal conditional policy.
We include two baselines based on the Go-Explore strategy(~\cite{ecoffet2019go}) that has been proved efficient in the GCRL setting: MEGA(~\cite{pitis2020maximum}) and PEG(~\cite{hu2023planning}). A Go-Explore agent firstly uses its goal-conditioned policy $\pi^G$ to approach a sampled exploration-inducing goal command $g$, referred to as the Go-phase. In the Explore-phase, it activates an exploration policy $\pi^E$ to explore the environment from the terminal state of the Go-phase. In contrast, \tool{} improves the generalization of world models to facilitate effective real-world environment exploration. During training, \tool{} collects trajectories that navigate between two goal states sampled from its candidate subgoal set, essentially replacing the "Explore-phase" in "Go-Explore" with another "Go-phase". MEGA commands the agent to rarely seen states at the frontier by using kernel density estimates (KDE) of state densities and chooses low-density goals from the replay buffer. PEG selects goal commands to guide an agent's goal-conditioned policy toward states with the highest exploration potential given its current level of training. This potential is defined as the expected accumulated exploration reward during the Explore-phase. Similar to \tool{}, our baseline methods, named \textbf{PEG-G} and \textbf{MEGA-G}, augment \textbf{GC-Dreamer} with the PEG and MEGA Go-Explore strategies, respectively. In these methods, the replay buffer $D$ contains not only trajectories sampled by the GCRL policy $\pi_G$ commanded by environment goals but also exploratory trajectories sampled using the Go-Explore strategies. The exploration policy $\pi^E$ in \textbf{PEG-G} and \textbf{MEGA-G} is the Plan2Explore policy from \cite{sekar2020planning}, which encourages the agent to actively search for states that induce disparities among an ensemble of world models.
% 
% which is a good metric to find a good goal $g$ for Go-phase which maximize the expected accumulated exploration reward of the Explore-phase.

%We use them to generate goals for exploration in \textbf{GC-Dreamer} in place of totally relying on the challenging goals from environments, represented as \textbf{PEG-G} and \textbf{MEGA-G}. 

%Both of \textbf{PEG-G} and \textbf{MEGA-G} follow the Go-Explore (\cite{ecoffet2019go}) exploration strategy 
 
%In \tool{}, we generate two-direction trajectories by randomly choosing arbitrary pair of subgoals found by APS as the intermediate starting point and ultimate target. %In order to compare the APS with other subgoal finding method, we consider the \textbf{Time-sample Hindsight Waypoints Sampling Strategy} from \textbf{Hindsight Planner}(\cite{lai2020hindsight}) to find the subgoals in place of APS in \tool{}. 

We note that \tool{} and the baselines are all implemented based on the Dreamer framework as realized in \textbf{GC-Dreamer}\footnote{\tool{} is not tied to a specific world model architecture and can be applied to any model-based RL framework.}. 
%The difference within \tool{} and all baselines is how to sample trajectories in the real environment to construct the Replay Buffer.

\subsection{Results}

\begin{figure}[t] 
  \centering
    \subfigure[Ant Maze]{\includegraphics[width=0.25\textwidth]{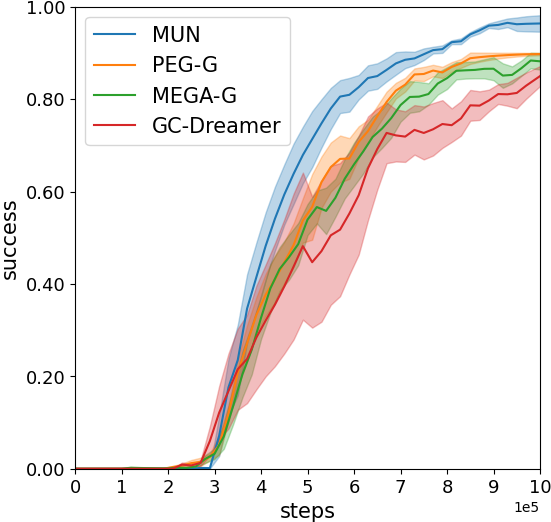}}
    \subfigure[Walker]{\includegraphics[width=0.25\textwidth]{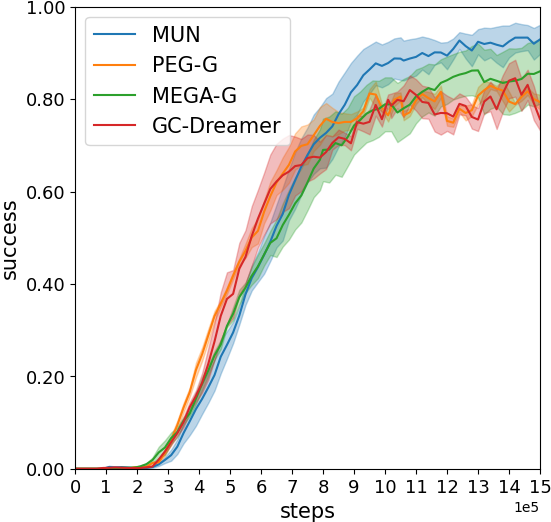}}
    \subfigure[3-Block Stacking]{\includegraphics[width=0.25\textwidth]{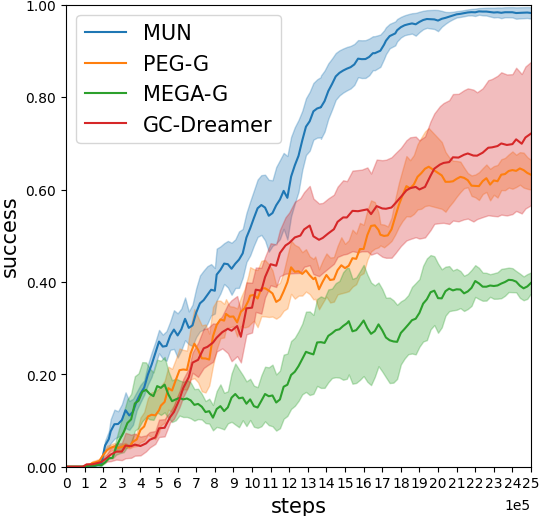}}
    \subfigure[Block Rotation]{\includegraphics[width=0.25\textwidth]{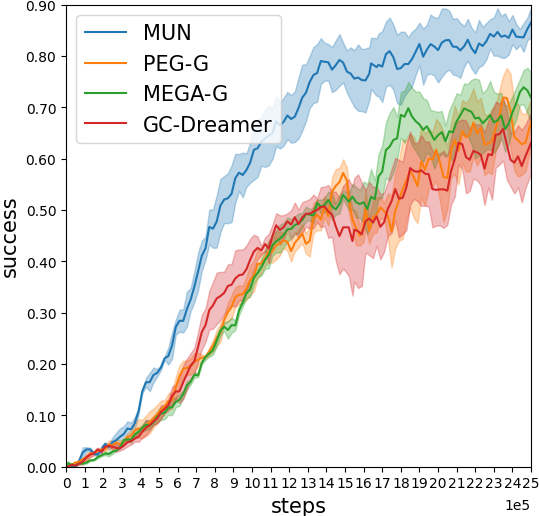}}
    \subfigure[Pen Rotation]{\includegraphics[width=0.25\textwidth]{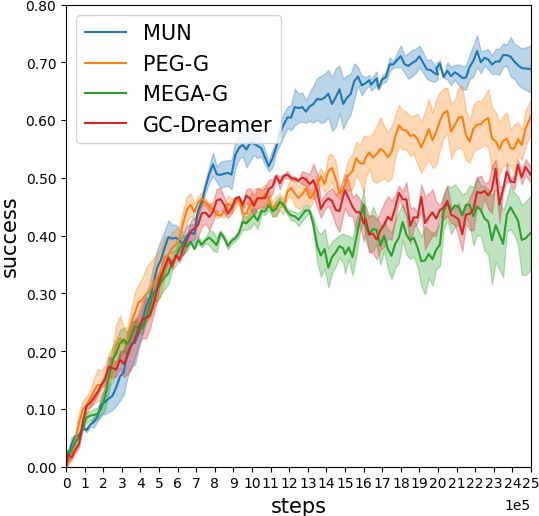}}
    \subfigure[Fetch Slide]{\includegraphics[width=0.25\textwidth]{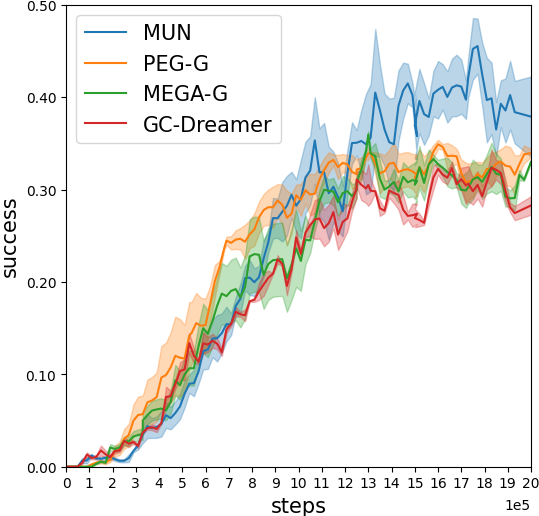}}
  \caption{Experiment results comparing \tool{} with the baselines over 5 random seeds.}
  \vspace{-10pt}
  \label{fig:exp results}
\end{figure}

Fig.~\ref{fig:exp results} shows the evaluation performance of \tool{} and all baselines across training. \tool{} demonstrates superior performance compared to the baseline models, excelling in both the final success rate and the speed of learning.
\tool{} outperforms the Go-Explore baselines (MEGA-G and PEG-G) across all tasks, demonstrating the effectiveness of the exploration strategy in \tool{} over the alternative Go-Explore strategies.
In the most challenging tasks—block stacking, block rotation, and pen rotation—\tool{} shows a significant margin of superiority. For example, \tool{} achieves over 95\% success rate on 3 block stacking, while all other baselines only manage to achieve around 60\% success rate on this task within 2.5M steps. MEGA-G and PEG-G heuristically pick exploration-inducing goals to initiate exploration by a separate policy. Since finding a goal state that is optimally aligned with both the goal-conditioned policy and the exploration policy is challenging, these methods can result in suboptimal goals, thereby slowing down exploration.
GC-Dreamer lacks a Go-Explore phase, which limits its exploration potential. Despite this, it can still perform comparably to or even better than MEGA-G and PEG-G in certain contexts.
This indicates that the Go-Explore strategy does not always guarantee improved exploration, and suboptimal goal-setting during the "Go-phase" can hinder exploration (see 3 block stacking). 

\begin{wrapfigure}[16]{r}{0.55\textwidth}
  \centering
    \subfigure[3-Block Stacking]{\includegraphics[width=0.25\textwidth]{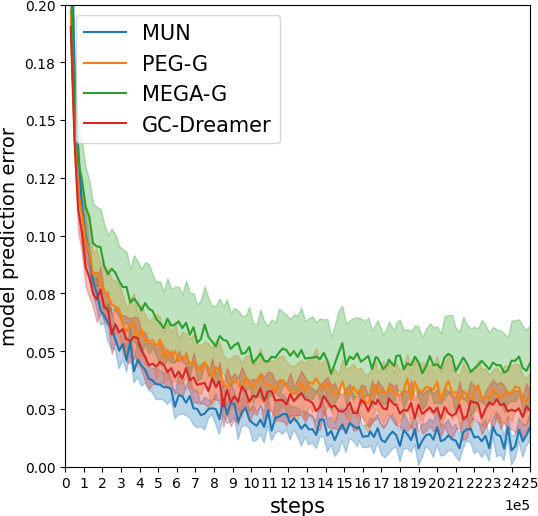}}
    \subfigure[Pen Rotation]{\includegraphics[width=0.25\textwidth]{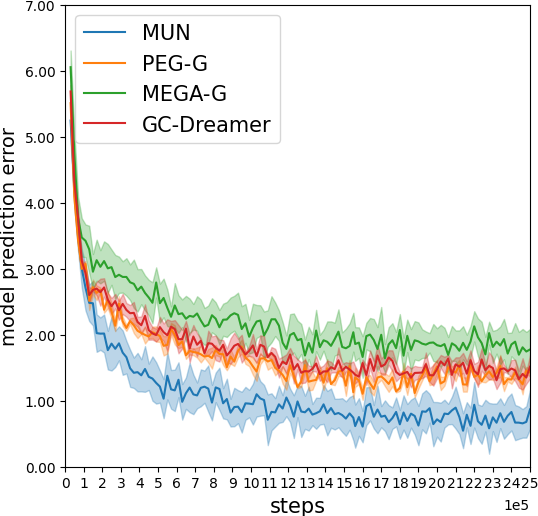}}
  \caption{The world model prediction error curves throughout the training steps for 3-Block Stacking and Pen Rotation.}
  %\vspace{-10pt}
  \label{fig: wm error curve}
\end{wrapfigure}

Fetch Slide is a non-prehensile manipulation task. This environment has asymmetric state transitions: when the puck is slid outside the robot's workspace, the manipulator cannot reach the puck's position to slide it backward due to physical constraints.
%Such asymmetric property seems to conflict with \tool{} because \tool{} emphasizes free transitions between different states.
MUN still outperforms the other baselines in this environment. We found MUN, with the DAD strategy, can discover key subgoals for this task, like contacting the puck, placing the manipulator at different angles around the puck, and stabilizing the manipulator upon reaching the goal (these key states result from distinct actions). MUN enables learning state transitions between these key subgoals to discover the high-level task structure. It learns a generalizable world model that handles sliding the puck between any positions within the workspace and predicts low probabilities for infeasible transitions from puck positions outside the workspace. Particularly, it enables the agent to hit the puck multiple times if it is within its workspace, thereby improving task success rates. That said, the current goal selection mechanism in \tool{} lacks a process to filter out infeasible goals from the current state, which could adversely affect sample efficiency. We left addressing this limitation and implementing a robust filtering mechanism for infeasible goals as a focus for future work.

We studied the prediction error of learned world models in \tool{} and the baselines. Fig.~\ref{fig: wm error curve} shows the one-step model prediction error throughout the training steps. %\tool{} achieves the lowest prediction error consistently across the training steps, demonstrating better world model quality. 
The world models trained by \tool{} show a much smaller generalization gap to the real environment compared to the baselines across the training steps. Consequently, \tool{} can effectively leverage these higher-quality world models to train policies that generalize better to the real environment. We present a quantitative comparison of the world model prediction quality between MUN and the baselines in terms of model prediction \emph{compounding error} in Appendix~\ref{subs: wm assessment}.

%for evaluation when generating the same length simulated trajectories. 

%More ablation experiment results will be discussed later to examine the significance of each component of \tool{}.

% The presented experimental results in Fig~\ref{fig:exp results} elucidate the efficacy of \tool{} and \toolhp{} across a diverse array of simulated environments, demonstrating their consistent superiority over the comparative baselines.
% In 3-Block Stacking, \tool{} and \toolhp{} exhibit a remarkable performance advantage, with \tool{} attaining a steady success rate over 90\% on hardest stacking 3 blocks goals and \toolhp{} achieving approximately 80\%.
% Compared to the surprising performance of \tool{} and \toolhp{}, other baselines only manage to achieve around 50\% success rate on this task within 2M steps. 
% This underscores the substantial advantage and efficiency of \tool{} in world model learning.
% For tasks involving intricate manipulation of robotic hand, such as manipulating block and pen, \tool{} is also capable of learning a world model that could simulate more comprehensive and complicated dynamic transformations. This enables it to achieve success rates surpassing various baselines. 
%indicating the predominant role of the Two-direction Replay Buffer in \tool{}. While the results in these three environments demonstrate the robustness of \tool{} under different Key Subgoal selection strategies, 

%

\subsection{Can DAD find key subgoals?}

\begin{figure}[h]
  \centering
  \includegraphics[width=0.9\textwidth]{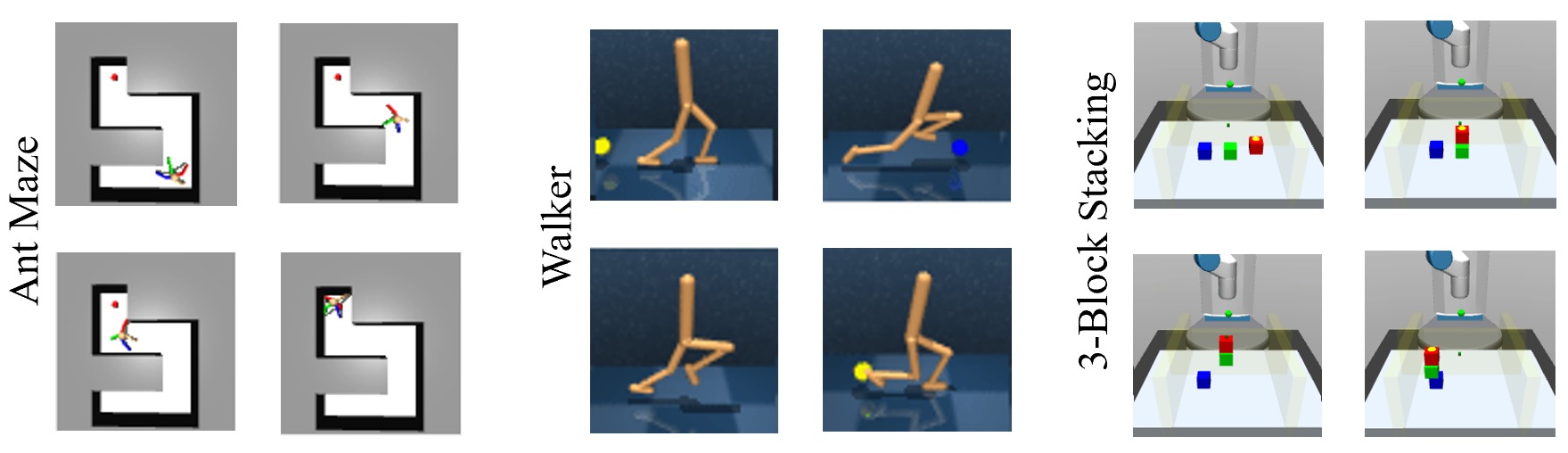}
  \caption{Key subgoals found by DAD (Algorithm~\ref{Algorithm2-APS}) in three environments: Ant-Maze, Walker, 3-Block Stacking. 
  They present the important landmarks on the path to the task goal regions.}
  \label{fig:subgoals}
\end{figure}

We visualize several subgoals found by the DAD algorithm during the training process in Fig.~\ref{fig:subgoals} for three environments: Ant-Maze, Walker, 3-Block Stacking.
In \textbf{Walker}, DAD successfully identifies the crucial joint angles and forces of the Walker robot during its forward locomotion, including standing, striding, jumping, landing, and leg support. 
In \textbf{Ant-Maze}, DAD recognizes significant motion variations at corridor corners. %Consequently, it successfully navigates between different rooms by identifying and traveling between such subgoals.
In \textbf{3-Block Stacking}, DAD successfully identifies crucial state transitions required during the stacking process.
%of three blocks by leveraging the diversity among actions  and recognizing the set of subgoals corresponding to actions with maximal dissimilarity. 
These critical subgoals include block grasping, lifting, horizontal movement, vertical movement, and gripper release.
%Based on the above example of APS successfully discovering subgoals, we can conclude that APS can efficiently, straightforwardly, and stably identifies the subgoal states required for completing tasks.
For more discussion about subgoals found by the DAD in other environments, please refer to Appendix~\ref{subs: subgoalsDAD}.

\subsection{Can \tool{} navigate between arbitrary subgoals?}

\begin{figure}[h] 
  \centering
  \includegraphics[width=0.9\textwidth]{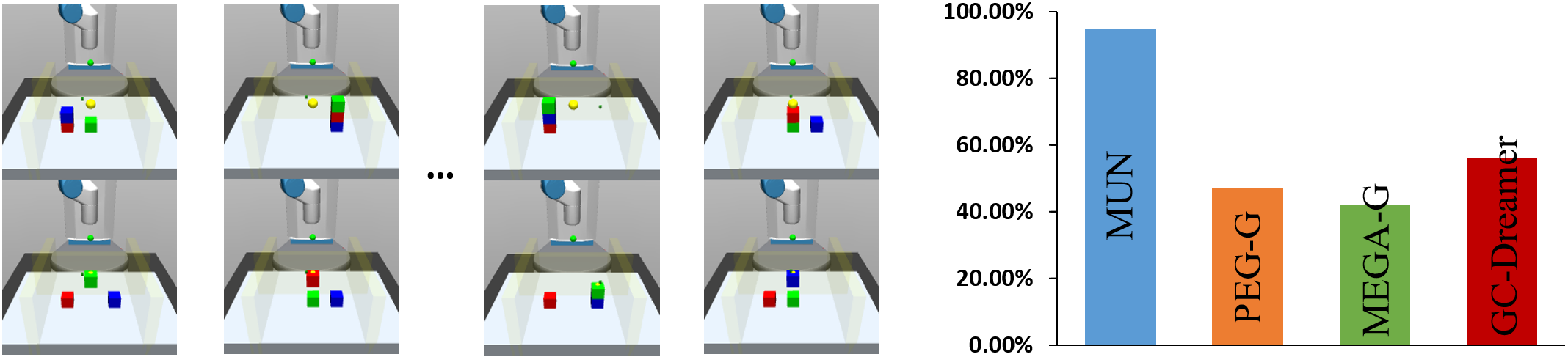}
  \caption{Experiment setup and results of navigation between any pair of subgoals in the 3-Block Stacking environment. 
  In the left part, the bottom section of each image depicts the ultimate evaluation goal for one evaluation episode, while the top section illustrates the manually set initial state.
  The right part shows the evaluation success rates.}
  \label{fig: navigate experiments}
  \end{figure}

%We also conduct the experiments to evaluate if \tool{} can reach unconstrained goals or navigates between any subgoals. 
As \tool{} is capable of identifying pivotal subgoal states necessary for complex tasks and training world models and policies for seamless transitions between these subgoals, we investigate MUN's capacity to generalize to new task settings concerning important subgoals. We set the initial state of the agent at one random subgoal and command it to reach another random subgoal. Such task setting is \emph{not provided to the agent during training.}
For the 3 Block Stacking task, we employ a set of 15 manually created subgoals representing various critical states in the block-stacking process, resulting in 225 unique combinations of initial states and test goals for evaluation. Each combination undergoes 10 repeated evaluations, totaling 2250 evaluation trajectories. These evaluations encompass both the forward and reverse processes of stacking and unstacking blocks, assessing the agent's proficiency in both task completion and restoration. For example, in the left portion of Fig.~\ref{fig: navigate experiments}, we visualize some subgoals used as initial task state in the upper part and some subgoals used as evaluation test goals in the lower part. The right section of Fig.~\ref{fig: navigate experiments} illustrates \tool{}'s superiority over the other baselines in these evaluation experiments, achieving the highest success rate through its ability to develop a robust and adaptable world model that generalizes to novel tasks. Additional results in different environments are provided in Appendix~\ref{subs: more navigation experiments}.
%to enabling seamless navigation between key subgoal states.

%Such evaluation include the forward and reverse processes of stacking blocks, assessing whether an agent, while capable of stacking blocks, is also able to unstack blocks to arbitary states. 
%We evaluate the comprehensive learning level of the World Model and the ability of the policy to reach goals in both forward task completion and reverse restoration paths.

%We want to evaluate if \tool{} perform well to achieve a random subgoal when the agent is initially placed at a arbitrary subgoal state.

%We can also assess the extent of the World Model's learning of the real world by observing the completion ability of navigation between arbitrary subgoal states by the agent. If the World Model's learning of the real world is not comprehensive enough, then policies trained using this World Model are likely to fail in evaluations involving such unrestricted start and end points. Conversely, if the World Model can learn the dynamic transformations and encoding between arbitrary sub-goals, then policies trained using it will perform well in such evaluations. This type of assessment represents an innovative evaluation approach in model based and goal-conditioned reinforcement learning, aiming to evaluate the comprehensive learning level of the World Model and the ability of the policy to reach goals in both forward task completion and reverse restoration paths.

\subsection{Ablation study}

We conducted the following ablation studies to investigate \tool{}'s exploration goal selection mechanism.
First, we investigated the effect of the number of subgoal states ($N_s$) in our algorithm. \tool{} sequentially traverses $N_s = 2$ goal states sampled from the replay buffer to explore the environment during each training episode. 
We introduced an ablation \textbf{MUN-Ns-3} that sets $N_s = 3$. This ablation aims to investigate whether increasing $N_s$ leads to improved learning performance.
Second, we considered an ablated version of \tool{}, named \textbf{\toolhp{}}, which replaces the goal sampling strategy DAD (Algorithm~\ref{Algorithm2-APS}) with a simple method that chooses goal states with fixed time interval in trajectories sampled from the replay buffer. This ablation investigates the importance of identifying key subgoal states, which represent pivotal milestones necessary to complete a complex task. It seeks to determine whether training world models from state transitions between these key states in \tool{} is essential, or if using any states from the replay buffer would suffice.
Lastly, we explored an alternative key subgoal discovery strategy. \tool{} identifies key subgoals for exploration as states in the replay buffer that result in distinct actions within the action space. We introduced an ablation, \textbf{MUN-KeyObs}, which directly discovers key subgoals from the state space by identifying centroids of (latent) state clusters in the replay buffer, following the strategy in \cite{zhang2021world}.

\begin{figure}[t] 
  \centering
    \subfigure[3-Block Stacking]{\includegraphics[width=0.3\textwidth]{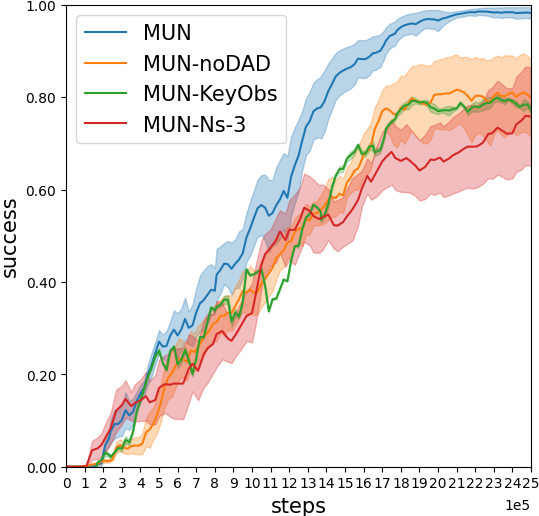}}
    \subfigure[Block Rotation]{\includegraphics[width=0.3\textwidth]{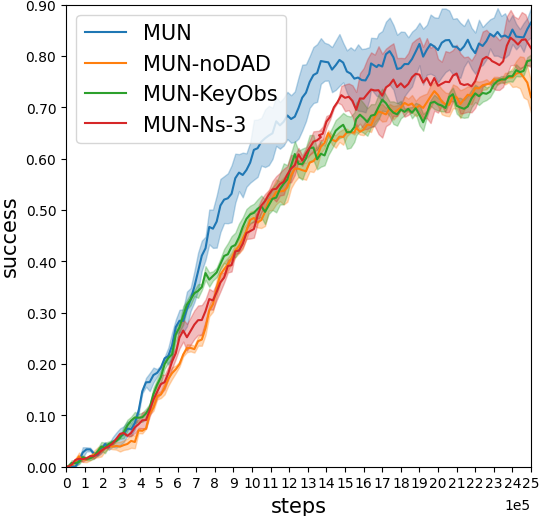}}
    \subfigure[Pen Rotation]{\includegraphics[width=0.3\textwidth]{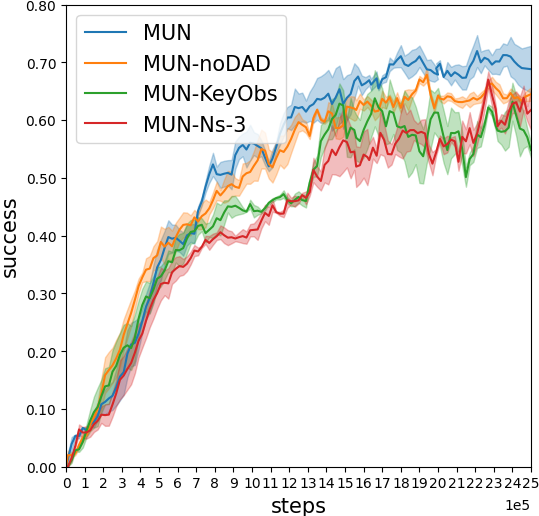}}
  \caption{Experiment results comparing \tool{} with its ablations over 5 random seeds.}
  \vspace{-10pt}
  \label{fig: ablation results}
\end{figure}

The results are depicted in Fig.~\ref{fig: ablation results}. \tool{} outperforms all ablated versions. Setting $N_s = 3$ slows down the training performance, supporting our claim it suffices to set $N_s = 2$.
%Comparing \tool{} and MUN-Ns-3, we can see that increasing $N_s$ beyond 2 does not necessarily lead to better performance. In fact, a larger number of subgoals can introduce unnecessary inefficiencies, either by slowing down the training process or leading to insufficient reaching of each goal state.
The performance of \toolhp{} and MUN-KeyObs does not match MUN, especially in the 3 Block Stacking environment, highlighting that discovering key subgoals in the action space (the DAD strategy) indeed contributes to higher performance and efficiency. 
%It can be seen that the ability of DAD to discover key subgoal states improves \tool{}'s generalizability to new tasks in this evaluation to different degrees in all environments compared to results of \toolhp{} and MUN-KeyObs, demonstrating that identifying key subgoal states by DAD for complex tasks indeed contributes to higher performance and efficiency. 
%It is noteworthy that all ablations achieve comparable success rates to \tool{} in these three tasks, while other baselines perform badly in such hard tasks. This suggests that \tool{}'s approach to learning a world model from state transitions between any states in the replay buffer (whether tracing back along recorded trajectories or transitioning across separate trajectories) is effective in bridging the generalization gap between the model and the real environment.
It is noteworthy that the ablation methods achieve a relatively small gap in success rates compared to MUN in the challenging Block Rotation and Pen Rotation environments.
This suggests that \tool{}'s approach to learning a world model from state transitions between any states in the replay buffer (whether tracing back along recorded trajectories or transitioning across separate trajectories) alone is effective in bridging the generalization gap between the model and the real environment. 
%The results from the other environments (3 Block Stacking, 275 Block Rotation, and Walker) demonstrate that identifying key subgoal states for complex tasks indeed 276 contributes to higher performance and efficiency

\section{Related Work}

Model-based reinforcement learning (MBRL) is a promising approach to reinforcement learning that learns a model of the environment and uses it to plan actions(\cite{sutton1991dyna,deisenroth2011pilco,oh2017value,chua2018deep}).
It has achieved remarkable success in numerous control tasks and games, such as chess(\cite{silver2017mastering,schrittwieser2020mastering,xu2022feasibility}), Atari games(\cite{hafner2020mastering,schrittwieser2020mastering,oh2017value}), 
continuous control tasks(\cite{kurutach2018model,buckman2018sample,hafner2019learning,janner2019trust}), and robotic manipulation tasks(\cite{lowrey2018plan, luo2018algorithmic}).
The dynamic model serves as a pivotal component of model-based reinforcement learning, primarily fulfilling two key roles: 
planning actions(\cite{deisenroth2011pilco,oh2017value,chua2018deep,lowrey2018plan,hafner2019learning}) 
or generating synthetic data to aid in the training of model-free reinforcement learning algorithms(\cite{janner2019trust,hafner2020mastering,hafner2023mastering}).
The primary drawback of the former lies in the excessive cost associated with long-term planning. To address this issue, the concept of ensemble has been employed to enhance performance(\cite{chua2018deep,kurutach2018model,buckman2018sample}).
\cite{oh2017value, hansen2022temporal} integrate the dynamics model with a value prediction network to improve the accuracy of long-term planning.
The latter also suffers from the potential bias of the model, which can result in inaccuracies in the generated data, thereby directly impacting policy learning(\cite{luo2018algorithmic, lai2021effective}).

Multi-goal reinforcement learning (RL) agents (\cite{schaul2015universal, plappert2018multi, ghosh2019learning}) acquire goal-conditioned behaviors capable of achieving and generalizing across diverse sets of objectives.
Researchers have been continuously exploring the integration of Model-based RL and Goal-conditioned RL(\cite{mendonca2021discovering,nair2020goal,zhang2020automatic}), 
leveraging the capabilities of dynamic models in planning and generating synthetic data to enhance the training efficiency and generalization of GCRL.
However, compared to traditional RL problems, GCRL faces more severe challenges regarding reward sparsity and exploration difficulties(\cite{ren2019exploration, florensa2018automatic, trott2019keeping}). 
These challenges often lead to significant biases in the learned World Model, consequently impairing the performance of goal-conditioned policy(\cite{mendonca2021discovering,hu2023planning}).
\cite{pong2019skew} propose to learn a maximum-entropy goal distribution, \cite{pitis2020maximum} encourage the agent to explore goals with low frequency of occurrence in the replay buffer,
\cite{sekar2020planning} introduce a planning algorithm to pick goals for exploration using world model.

The World Model holds inherent advantages for GCRL, as it often enables faster exploration and facilitates the 
training of a more generalized policy(\cite{mccarthy2021imaginary, shyam2019model, hu2023planning, sekar2020planning}).
However, within the GCRL framework, learning a reliable World Model is a crucial prerequisite for developing excellent policies(\cite{zhang2024storm,young2022benefits,wang2023live,lai2021effective}). 
\cite{kauvar2023curious} propose a curiosity-driven exploration method, which is focused on replay buffer management. 
\cite{hansen2022modem} use demonstration data as a supplement to the replay buffer to learn a more reliable World Model.
Previous work has often focused on devising more appropriate objectives when sampling real trajectory data from the environment to 
enrich the diversity of dynamic transitions in the replay buffer(\cite{nair2020goal,charlesworth2020plangan,trott2019keeping,florensa2018automatic,campero2020learning}). 
However, they overlooked the overall direction of dynamic transitions within the data which extremely affects the richness of dynamic transitions to learn a comprehensive World Model.

\section{Conclusion}

In summary, we introduce \tool{}, a novel goal-directed exploration algorithm designed for effective world modeling of seamless transitions between arbitrary states in replay buffers, whether retracing along recorded trajectories or transitioning between states on separate trajectories. As the quality of the world model improves, \tool{} demonstrates high efficacy in learning goal-conditioned policies in sparse-reward environments. Additionally, we present a practical strategy DAD for identifying pivotal subgoal states, which act as critical milestones in completing complex tasks. %Through training world models for seamless transitions between these milestones, \tool{} enables the learning of policies capable of adapting to novel goal scenarios.
The experimental results underscored the effectiveness of \tool{} in strengthening the reliability of world models and learning policies capable of adapting to novel test goals.

%We proposes \tool{}, a novel goal-directed exploration algorithm designed to address the challenges of generalizing learned world models. By modeling state transitions between arbitrary subgoal states in the replay buffer, \tool{} facilitates the learning of policies capable of navigating between key states essential for task completion. Our 

%an approach to learn a comprehensive and robust world model. \tool{} samples trajectories transiting from and to any key subgoal states to collect a richer dynamic replay buffer to train world model. 
%We have also defined a concise and efficient method called APS to discover key subgoals. 
%The core idea of \tool{} and APS can also be applied in model-free RL framework to construct a Replay Buffer with more comprehensive and richer dynamic transitions, thus enhancing policy capabilities. 
%However, the question of whether there exist more accurate and efficient methods to replace APS in discovering key subgoal states remains an open area for research, and our work establishes a paradigm for addressing this challenge.

\section*{Reproducibility Statement} 

The code of \tool{} is provided on \hyperlink{https://github.com/RU-Automated-Reasoning-Group/MUN}{https://github.com/RU-Automated-Reasoning-Group/MUN}. For hyperparameter settings and baseline pseudocode, please refer to Appendix~\ref{supp:baselines} and Appendix~\ref{subsec: hyperparameters}.

\section*{Acknowledgements}

We thank the anonymous reviewers for their comments and suggestions. This work was supported by NSF Award \#CCF-2124155.

\bibliography{Reference/Reference}

\begin{thebibliography}{}

\bibitem[Andrychowicz et~al., 2017]{andrychowicz2017hindsight}
Andrychowicz, M., Wolski, F., Ray, A., Schneider, J., Fong, R., Welinder, P., McGrew, B., Tobin, J., Pieter~Abbeel, O., and Zaremba, W. (2017).
\newblock Hindsight experience replay.
\newblock {\em Advances in neural information processing systems}, 30.

\bibitem[Bellemare et~al., 2016]{bellemare2016unifying}
Bellemare, M., Srinivasan, S., Ostrovski, G., Schaul, T., Saxton, D., and Munos, R. (2016).
\newblock Unifying count-based exploration and intrinsic motivation.
\newblock {\em Advances in neural information processing systems}, 29.

\bibitem[Buckman et~al., 2018]{buckman2018sample}
Buckman, J., Hafner, D., Tucker, G., Brevdo, E., and Lee, H. (2018).
\newblock Sample-efficient reinforcement learning with stochastic ensemble value expansion.
\newblock {\em Advances in neural information processing systems}, 31.

\bibitem[Burda et~al., 2018]{burda2018exploration}
Burda, Y., Edwards, H., Storkey, A., and Klimov, O. (2018).
\newblock Exploration by random network distillation.
\newblock {\em arXiv preprint arXiv:1810.12894}.

\bibitem[Campero et~al., 2020]{campero2020learning}
Campero, A., Raileanu, R., K{\"u}ttler, H., Tenenbaum, J.~B., Rockt{\"a}schel, T., and Grefenstette, E. (2020).
\newblock Learning with amigo: Adversarially motivated intrinsic goals.
\newblock {\em arXiv preprint arXiv:2006.12122}.

\bibitem[Charlesworth and Montana, 2020]{charlesworth2020plangan}
Charlesworth, H. and Montana, G. (2020).
\newblock Plangan: Model-based planning with sparse rewards and multiple goals.
\newblock {\em Advances in Neural Information Processing Systems}, 33:8532--8542.

\bibitem[Chua et~al., 2018]{chua2018deep}
Chua, K., Calandra, R., McAllister, R., and Levine, S. (2018).
\newblock Deep reinforcement learning in a handful of trials using probabilistic dynamics models.
\newblock {\em Advances in neural information processing systems}, 31.

\bibitem[Deisenroth and Rasmussen, 2011]{deisenroth2011pilco}
Deisenroth, M. and Rasmussen, C.~E. (2011).
\newblock Pilco: A model-based and data-efficient approach to policy search.
\newblock In {\em Proceedings of the 28th International Conference on machine learning (ICML-11)}, pages 465--472.

\bibitem[Ecoffet et~al., 2019]{ecoffet2019go}
Ecoffet, A., Huizinga, J., Lehman, J., Stanley, K.~O., and Clune, J. (2019).
\newblock Go-explore: a new approach for hard-exploration problems.
\newblock {\em arXiv preprint arXiv:1901.10995}.

\bibitem[Eldar et~al., 1997]{eldar1997farthest}
Eldar, Y., Lindenbaum, M., Porat, M., and Zeevi, Y.~Y. (1997).
\newblock The farthest point strategy for progressive image sampling.
\newblock {\em IEEE transactions on image processing}, 6(9):1305--1315.

\bibitem[Florensa et~al., 2018]{florensa2018automatic}
Florensa, C., Held, D., Geng, X., and Abbeel, P. (2018).
\newblock Automatic goal generation for reinforcement learning agents.
\newblock In {\em International conference on machine learning}, pages 1515--1528. PMLR.

\bibitem[Georgiev et~al., 2024]{georgiev2024pwm}
Georgiev, I., Giridhar, V., Hansen, N., and Garg, A. (2024).
\newblock Pwm: Policy learning with large world models.
\newblock {\em arXiv preprint arXiv:2407.02466}.

\bibitem[Ghosh et~al., 2019]{ghosh2019learning}
Ghosh, D., Gupta, A., Reddy, A., Fu, J., Devin, C., Eysenbach, B., and Levine, S. (2019).
\newblock Learning to reach goals via iterated supervised learning.
\newblock {\em arXiv preprint arXiv:1912.06088}.

\bibitem[Hafner et~al., 2019a]{hafner2019dream}
Hafner, D., Lillicrap, T., Ba, J., and Norouzi, M. (2019a).
\newblock Dream to control: Learning behaviors by latent imagination.
\newblock {\em arXiv preprint arXiv:1912.01603}.

\bibitem[Hafner et~al., 2019b]{hafner2019learning}
Hafner, D., Lillicrap, T., Fischer, I., Villegas, R., Ha, D., Lee, H., and Davidson, J. (2019b).
\newblock Learning latent dynamics for planning from pixels.
\newblock In {\em International conference on machine learning}, pages 2555--2565. PMLR.

\bibitem[Hafner et~al., 2020]{hafner2020mastering}
Hafner, D., Lillicrap, T., Norouzi, M., and Ba, J. (2020).
\newblock Mastering atari with discrete world models.
\newblock {\em arXiv preprint arXiv:2010.02193}.

\bibitem[Hafner et~al., 2023]{hafner2023mastering}
Hafner, D., Pasukonis, J., Ba, J., and Lillicrap, T. (2023).
\newblock Mastering diverse domains through world models.
\newblock {\em arXiv preprint arXiv:2301.04104}.

\bibitem[Hansen et~al., 2022a]{hansen2022modem}
Hansen, N., Lin, Y., Su, H., Wang, X., Kumar, V., and Rajeswaran, A. (2022a).
\newblock Modem: Accelerating visual model-based reinforcement learning with demonstrations.
\newblock {\em arXiv preprint arXiv:2212.05698}.

\bibitem[Hansen et~al., 2023]{hansen2023td}
Hansen, N., Su, H., and Wang, X. (2023).
\newblock Td-mpc2: Scalable, robust world models for continuous control.
\newblock {\em arXiv preprint arXiv:2310.16828}.

\bibitem[Hansen et~al., 2022b]{hansen2022temporal}
Hansen, N., Wang, X., and Su, H. (2022b).
\newblock Temporal difference learning for model predictive control.
\newblock {\em arXiv preprint arXiv:2203.04955}.

\bibitem[Hu et~al., 2023]{hu2023planning}
Hu, E.~S., Chang, R., Rybkin, O., and Jayaraman, D. (2023).
\newblock Planning goals for exploration.
\newblock {\em arXiv preprint arXiv:2303.13002}.

\bibitem[Janner et~al., 2019]{janner2019trust}
Janner, M., Fu, J., Zhang, M., and Levine, S. (2019).
\newblock When to trust your model: Model-based policy optimization.
\newblock {\em Advances in neural information processing systems}, 32.

\bibitem[Kauvar et~al., 2023]{kauvar2023curious}
Kauvar, I., Doyle, C., Zhou, L., and Haber, N. (2023).
\newblock Curious replay for model-based adaptation.
\newblock {\em arXiv preprint arXiv:2306.15934}.

\bibitem[Kurutach et~al., 2018]{kurutach2018model}
Kurutach, T., Clavera, I., Duan, Y., Tamar, A., and Abbeel, P. (2018).
\newblock Model-ensemble trust-region policy optimization.
\newblock {\em arXiv preprint arXiv:1802.10592}.

\bibitem[Lai et~al., 2021]{lai2021effective}
Lai, H., Shen, J., Zhang, W., Huang, Y., Zhang, X., Tang, R., Yu, Y., and Li, Z. (2021).
\newblock On effective scheduling of model-based reinforcement learning.
\newblock {\em Advances in Neural Information Processing Systems}, 34:3694--3705.

\bibitem[Lai et~al., 2020]{lai2020hindsight}
Lai, Y., Wang, W., Yang, Y., Zhu, J., and Kuang, M. (2020).
\newblock Hindsight planner.
\newblock In {\em Proceedings of the 19th International Conference on Autonomous Agents and MultiAgent Systems}, pages 690--698.

\bibitem[Lanier, 2019]{lanier2019curiosity}
Lanier, J.~B. (2019).
\newblock {\em Curiosity-driven multi-criteria hindsight experience replay}.
\newblock University of California, Irvine.

\bibitem[Li et~al., 2020]{li2020towards}
Li, R., Jabri, A., Darrell, T., and Agrawal, P. (2020).
\newblock Towards practical multi-object manipulation using relational reinforcement learning.
\newblock In {\em 2020 ieee international conference on robotics and automation (icra)}, pages 4051--4058. IEEE.

\bibitem[Lowrey et~al., 2018]{lowrey2018plan}
Lowrey, K., Rajeswaran, A., Kakade, S., Todorov, E., and Mordatch, I. (2018).
\newblock Plan online, learn offline: Efficient learning and exploration via model-based control.
\newblock {\em arXiv preprint arXiv:1811.01848}.

\bibitem[Luo et~al., 2018]{luo2018algorithmic}
Luo, Y., Xu, H., Li, Y., Tian, Y., Darrell, T., and Ma, T. (2018).
\newblock Algorithmic framework for model-based deep reinforcement learning with theoretical guarantees.
\newblock {\em arXiv preprint arXiv:1807.03858}.

\bibitem[McCarthy et~al., 2021]{mccarthy2021imaginary}
McCarthy, R., Wang, Q., and Redmond, S.~J. (2021).
\newblock Imaginary hindsight experience replay: Curious model-based learning for sparse reward tasks.
\newblock {\em arXiv preprint arXiv:2110.02414}.

\bibitem[Mendonca et~al., 2021]{mendonca2021discovering}
Mendonca, R., Rybkin, O., Daniilidis, K., Hafner, D., and Pathak, D. (2021).
\newblock Discovering and achieving goals via world models.
\newblock {\em Advances in Neural Information Processing Systems}, 34:24379--24391.

\bibitem[Nagabandi et~al., 2020]{nagabandi2020deep}
Nagabandi, A., Konolige, K., Levine, S., and Kumar, V. (2020).
\newblock Deep dynamics models for learning dexterous manipulation.
\newblock In {\em Conference on Robot Learning}, pages 1101--1112. PMLR.

\bibitem[Nair et~al., 2018]{nair2018overcoming}
Nair, A., McGrew, B., Andrychowicz, M., Zaremba, W., and Abbeel, P. (2018).
\newblock Overcoming exploration in reinforcement learning with demonstrations.
\newblock In {\em 2018 IEEE international conference on robotics and automation (ICRA)}, pages 6292--6299. IEEE.

\bibitem[Nair et~al., 2020]{nair2020goal}
Nair, S., Savarese, S., and Finn, C. (2020).
\newblock Goal-aware prediction: Learning to model what matters.
\newblock In {\em International Conference on Machine Learning}, pages 7207--7219. PMLR.

\bibitem[Oh et~al., 2017]{oh2017value}
Oh, J., Singh, S., and Lee, H. (2017).
\newblock Value prediction network.
\newblock {\em Advances in neural information processing systems}, 30.

\bibitem[Pathak et~al., 2017]{pathak2017curiosity}
Pathak, D., Agrawal, P., Efros, A.~A., and Darrell, T. (2017).
\newblock Curiosity-driven exploration by self-supervised prediction.
\newblock In {\em International conference on machine learning}, pages 2778--2787. PMLR.

\bibitem[Paul et~al., 2019]{paul2019learning}
Paul, S., Vanbaar, J., and Roy-Chowdhury, A. (2019).
\newblock Learning from trajectories via subgoal discovery.
\newblock {\em Advances in Neural Information Processing Systems}, 32.

\bibitem[Pislar et~al., 2021]{pislar2021should}
Pislar, M., Szepesvari, D., Ostrovski, G., Borsa, D., and Schaul, T. (2021).
\newblock When should agents explore?
\newblock {\em arXiv preprint arXiv:2108.11811}.

\bibitem[Pitis et~al., 2020]{pitis2020maximum}
Pitis, S., Chan, H., Zhao, S., Stadie, B., and Ba, J. (2020).
\newblock Maximum entropy gain exploration for long horizon multi-goal reinforcement learning.
\newblock In {\em International Conference on Machine Learning}, pages 7750--7761. PMLR.

\bibitem[Plappert et~al., 2018]{plappert2018multi}
Plappert, M., Andrychowicz, M., Ray, A., McGrew, B., Baker, B., Powell, G., Schneider, J., Tobin, J., Chociej, M., Welinder, P., et~al. (2018).
\newblock Multi-goal reinforcement learning: Challenging robotics environments and request for research.
\newblock {\em arXiv preprint arXiv:1802.09464}.

\bibitem[Pong et~al., 2019]{pong2019skew}
Pong, V.~H., Dalal, M., Lin, S., Nair, A., Bahl, S., and Levine, S. (2019).
\newblock Skew-fit: State-covering self-supervised reinforcement learning.
\newblock {\em arXiv preprint arXiv:1903.03698}.

\bibitem[Ren et~al., 2019]{ren2019exploration}
Ren, Z., Dong, K., Zhou, Y., Liu, Q., and Peng, J. (2019).
\newblock Exploration via hindsight goal generation.
\newblock {\em Advances in Neural Information Processing Systems}, 32.

\bibitem[Schaul et~al., 2015]{schaul2015universal}
Schaul, T., Horgan, D., Gregor, K., and Silver, D. (2015).
\newblock Universal value function approximators.
\newblock In {\em International conference on machine learning}, pages 1312--1320. PMLR.

\bibitem[Schrittwieser et~al., 2020]{schrittwieser2020mastering}
Schrittwieser, J., Antonoglou, I., Hubert, T., Simonyan, K., Sifre, L., Schmitt, S., Guez, A., Lockhart, E., Hassabis, D., Graepel, T., et~al. (2020).
\newblock Mastering atari, go, chess and shogi by planning with a learned model.
\newblock {\em Nature}, 588(7839):604--609.

\bibitem[Sekar et~al., 2020]{sekar2020planning}
Sekar, R., Rybkin, O., Daniilidis, K., Abbeel, P., Hafner, D., and Pathak, D. (2020).
\newblock Planning to explore via self-supervised world models.
\newblock In {\em International conference on machine learning}, pages 8583--8592. PMLR.

\bibitem[Shyam et~al., 2019]{shyam2019model}
Shyam, P., Ja{\'s}kowski, W., and Gomez, F. (2019).
\newblock Model-based active exploration.
\newblock In {\em International conference on machine learning}, pages 5779--5788. PMLR.

\bibitem[Silver et~al., 2017]{silver2017mastering}
Silver, D., Hubert, T., Schrittwieser, J., Antonoglou, I., Lai, M., Guez, A., Lanctot, M., Sifre, L., Kumaran, D., Graepel, T., et~al. (2017).
\newblock Mastering chess and shogi by self-play with a general reinforcement learning algorithm.
\newblock {\em arXiv preprint arXiv:1712.01815}.

\bibitem[Sutton, 1991]{sutton1991dyna}
Sutton, R.~S. (1991).
\newblock Dyna, an integrated architecture for learning, planning, and reacting.
\newblock {\em ACM Sigart Bulletin}, 2(4):160--163.

\bibitem[Trott et~al., 2019]{trott2019keeping}
Trott, A., Zheng, S., Xiong, C., and Socher, R. (2019).
\newblock Keeping your distance: Solving sparse reward tasks using self-balancing shaped rewards.
\newblock {\em Advances in Neural Information Processing Systems}, 32.

\bibitem[Tuyls et~al., 2022]{tuyls2022multi}
Tuyls, J., Yao, S., Kakade, S., and Narasimhan, K. (2022).
\newblock Multi-stage episodic control for strategic exploration in text games.
\newblock {\em arXiv preprint arXiv:2201.01251}.

\bibitem[Wang et~al., 2023]{wang2023live}
Wang, X., Wongkamjan, W., Jia, R., and Huang, F. (2023).
\newblock Live in the moment: Learning dynamics model adapted to evolving policy.
\newblock In {\em International Conference on Machine Learning}, pages 36470--36493. PMLR.

\bibitem[Williams et~al., 2015]{williams2015model}
Williams, G., Aldrich, A., and Theodorou, E. (2015).
\newblock Model predictive path integral control using covariance variable importance sampling.
\newblock {\em arXiv preprint arXiv:1509.01149}.

\bibitem[Xu et~al., 2022]{xu2022feasibility}
Xu, Y., Hansen, N., Wang, Z., Chan, Y.-C., Su, H., and Tu, Z. (2022).
\newblock On the feasibility of cross-task transfer with model-based reinforcement learning.
\newblock {\em arXiv preprint arXiv:2210.10763}.

\bibitem[Young et~al., 2022]{young2022benefits}
Young, K., Ramesh, A., Kirsch, L., and Schmidhuber, J. (2022).
\newblock The benefits of model-based generalization in reinforcement learning.
\newblock {\em arXiv preprint arXiv:2211.02222}.

\bibitem[Yu et~al., 2020]{yu2020mopo}
Yu, T., Thomas, G., Yu, L., Ermon, S., Zou, J.~Y., Levine, S., Finn, C., and Ma, T. (2020).
\newblock Mopo: Model-based offline policy optimization.
\newblock {\em Advances in Neural Information Processing Systems}, 33:14129--14142.

\bibitem[Zhang et~al., 2021]{zhang2021world}
Zhang, L., Yang, G., and Stadie, B.~C. (2021).
\newblock World model as a graph: Learning latent landmarks for planning.
\newblock In {\em International conference on machine learning}, pages 12611--12620. PMLR.

\bibitem[Zhang et~al., 2024]{zhang2024storm}
Zhang, W., Wang, G., Sun, J., Yuan, Y., and Huang, G. (2024).
\newblock Storm: Efficient stochastic transformer based world models for reinforcement learning.
\newblock {\em Advances in Neural Information Processing Systems}, 36.

\bibitem[Zhang et~al., 2020]{zhang2020automatic}
Zhang, Y., Abbeel, P., and Pinto, L. (2020).
\newblock Automatic curriculum learning through value disagreement.
\newblock {\em Advances in Neural Information Processing Systems}, 33:7648--7659.

\end{thebibliography}
\bibliographystyle{apalike}

%%%%%%%%%%%%%%%%%%%%%%%%%%%%%%%%%%%%%%%%%%%%%%%%%%%%%%%%%%%%

% Start the appendix

\newpage 

\appendix
\addcontentsline{toc}{section}{Appendix} 
{\fontsize{20}{14}\selectfont \textbf{Appendix}}
\parttoc 

\section{Extended Background}

\subsection{Dreamer World Model}
\label{subs: rssm}
The RSSM consists of an encoder, a recurrent model, a representation model, a transition predictor, and a decoder, as formulated in Equation~\ref{eq:world_model}.
And it employs an end-to-end training methodology, where its parameters are jointly optimized based on the loss functions of various components, 
including dynamic transition prediction, reward prediction, and observation encoding-decoding. 
These components often operate in a latent space rather than the original observation space, as encoded by the World Model. 
Therefore, during end-to-end training, the losses of all components indirectly optimize the latent space. 

The encoder $f_E$ encodes the input state $x_t$ into a embed state $e_t$, which is then fed with the deterministic state $h_t$ into the representation model $q_\phi$ to generate the posterior state $z_t$.
The transition predictor $p_\phi$ predicts the prior state $\hat{z}_t$ based on the deterministic state $h_t$ without access to the current input state $x_t$.
Using the concatenation of either $(h_{t}, z_{t})$ or $(h_{t}, \hat{z}_t)$ as input, the recurrent transition function $f{\phi}$ iteratively updates the deterministic state $h_t$ with given action $a_t$.

\begin{equation}\label{eq:world_model}
  \begin{aligned}
    \text{Encoder:\qquad} & e_t = f_E(e_t | x_t)\\
    \text{Recurrent model:\qquad} & h_t = f_\phi(h_{t-1}, z_{t-1}, a_{t-1}) \\
    \text{Representation model:\qquad} & z_t \sim q_\phi(z_t | h_t, e_t) \\
    \text{Transition predictor:\qquad} & \hat{z}_t \sim p_\phi(\hat{z}_t | h_t) \\
    \text{Decoder:\qquad} & \hat{x}_t \sim f_D(\hat{x}_t | h_t, z_t)
  \end{aligned}
\end{equation}

\subsection{Temporal Distance Training in LEXA}
\label{subs: Dt-training}

The goal-reaching reward $r^G$ is defined by the self-supervised temporal distance objective (\cite{mendonca2021discovering}) 
which aims to minimize the number of action steps needed to transition from the current state to a goal state within imagined rollouts.
We use $b_t$ to denote the concatenate of the deterministic state $h_t$ and the posterior state $z_t$ at time step $t$.

\begin{equation}\label{eq: temporal_distance_reward1}
  b_t = (h_t, z_t)
\end{equation}

The temporal distance $D_t$ is trained by sampling pairs of imagined states $b_t, b_{t+k}$ from imagined rollouts and predicting the action steps number $k$ between the embedding of them, 
with a predicted embedding $\hat{e}_t$ from $b_t$ to approximate the true embedding $e_t$ of the observation $x_t$.

\begin{equation}\label{eq: temporal_distance_reward2}
  \text{Predicted embedding:\qquad} emb(b_t) = \hat{e}_t \approx e_t, \qquad \text{where\quad} e_t = f_E(x_t)
\end{equation}

\begin{equation}\label{eq: temporal_distance_reward3}
  \text{Temporal distance: } D_t(\hat{e}_t, \hat{e}_{t+k}) \approx k/H \qquad \text{where\quad} \hat{e}_t = emb(b_t)\quad \hat{e}_{t+k} = emb(b_{t+k})
\end{equation}

\begin{equation}\label{eq: temporal_distance_reward4}
  r^G_t(b_t, b_{t+k}) = -D_t(\hat{e}_t, \hat{e}_{t+k})
\end{equation}

\section{Limitations and Future Work}

The \tool{} has provided powerful guidance in enhancing world model learning by repeatedly studying the transitions between various key states. 
This allows the acquisition of richer dynamic transitions and deepens the world model's understanding of the real world. 
However, such a framework requires an efficient strategy for discovering key states, as evidenced by the comparative results of the \tool{} and \toolhp{}. 
We found that although DAD excels in discovering key states with its simple and efficient method, 
it will identify ineffective and task-irrelevant states in tasks with highly complex action spaces or weak correlations between goal space and action space. 
This can lead to the degradation of the \tool{} architecture due to poor-quality subgoals, resulting in a substantial amount of ineffective sampling in the environment. 
Therefore, for environments with more complex action and goal spaces, we need to develop a more robust and effective method for discovering subgoals than DAD. 
Only with an efficient and powerful self-supervised subgoal discovery mechanism can the \tool{} framework be fully utilized.

Meanwhile, \tool{} autonomously discovers subgoals and learns a more robust and comprehensive world model by randomly navigating between subgoals. 
Although the \tool{} has achieved huge success in model-based reinforcement learning (MBRL), we believe it can also be applied to general model-free methods. 
General model-free methods do not require learning a world model and have a simpler architecture. 
The \tool{} can directly guide the goal-conditioned policy to enhance learning in navigation between different subgoals. 
It can use sampled trajectories to learn this policy directly, bypassing the use of the world model to train policies and value functions through simulated trajectories, thereby enhancing the agent's ability to reach unconstrained goals. 
Therefore, we plan to explore the application and effectiveness of the \tool{} in model-free RL in the future and develop a new robust self-supervised subgoal discovery mechanism to make the \tool{} applicable to more complex environments.

\section{Environments}
\label{sec: envs}

\subsection{3-Block Stacking} 
In this task, the robot must stack three blocks in different colors into a tower shape. 
While PEG assesses goals of varying difficulty levels: 3 easy goals (picking up a single block), 6 medium goals (stacking two blocks), and 6 hard goals (stacking three blocks),
our evaluation is focused solely on the 6 hard goals, and we use only 3 hard goals of them as the guiding goals from the training environment.
Training and evaluating with only the hardest goals imposes a significant challenge for the \tool{}. 
However, we observed that the \tool{} can spontaneously discover additional easy and medium goals through DAD, as these serve as critical transitional states toward the hard goals.
The environment is characterized by a 14-dimensional state and goal space. The first five dimensions capture the gripper's state, while the remaining nine dimensions correspond to the $xyz$ positions of each block. 
The action space is 4-dimensional, with three dimensions dedicated to the gripper's $xyz$ movements and the fourth dimension controlling the gripper's finger movement. 
Success is defined by achieving an L2 distance of less than 3 cm between each block's $xyz$ position and its target position. 
This environment is a modified version of the FetchStack3 environment from \cite{pitis2020maximum}, designed to better test the robot's precision in stacking.

\subsection{Walker}
In this environment, a 2D walker robot is trained and evaluated on its ability to move across a flat surface. 
The environment's implementation is based on the code from \cite{mendonca2021discovering}. 
To thoroughly assess the agent's capability and precision in covering longer distances, 
we expanded the evaluation goals to 12 ($\pm13, \pm16, \pm19, \pm22, \pm25, \pm28$) along the $x$ axis from the initial position.
In our training setup for the \tool{}, we only use the goals at $\pm13$ and $\pm16$ provided by the environment, but we evaluate the agent's performance across all 12 goals. 
Success is measured by verifying whether the agent's $x$ position is within a small margin of the target $x$ position.
The state and goal space in this environment are nine-dimensional, comprising the walker's $xz$ positions and its joint angles. 
This configuration ensures a comprehensive evaluation of the walker's locomotion capabilities.

\subsection{Ant Maze} 
This environment builds upon the Ant Maze from \cite{pitis2020maximum}, incorporating a few modifications. 
The state and goal spaces in the Ant Maze environment are highly complex, totaling 29 dimensions. These dimensions include the ant's $xyz$ position, joint angles, and velocities. The first three dimensions account for the $xyz$ position, the next 12 dimensions capture the joint angles of the ant's limbs, and the remaining 14 dimensions represent the velocities of the joints and the ant's movements in the $xy$ plane. The action space consists of 8 dimensions, controlling the hip and ankle movements of the ant's legs.We matched the goal space to the state space, which includes the ant's $xyz$ coordinates, joint positions, and velocities. We also introduced an additional room in the top left to increase the difficulty like PEG.
In this scenario, the ant robot must traverse from the bottom left to the top left of a maze, navigating through various corridors. The task is particularly challenging due to its lengthy duration—each episode lasts 500 timesteps—and the significant distance the ant must cover. Unlike PEG, which evaluates goals in both the central hallway and the top left room, our evaluation focuses exclusively on the four most difficult goals located in the top left room.
For training, we utilize all 32 goals throughout the maze. The maze itself measures about 6 by 8 meters. The ant succeeds if its $xy$ position is within 1.0 meter of the goal, roughly the size of a single cell in the maze.

\subsection{Fetch Slide}
In this task, a robotic arm with a two-fingered gripper must push an object along a flat surface to a specific goal location. We use the "FetchSlide-v1" environment from Gymnasium, where the robot operates in a 25-dimensional state space that includes the robot’s joint states, object position, and goal information. The goal space is 3-dimensional, representing the target coordinates for the object. Each episode presents a unique random goal location within a bounded area, requiring the agent to adjust its pushing strategy accordingly.
A key challenge in Fetch Slide lies in the indirect manipulation of the object. The agent must accurately control the force and direction of its push while accounting for physical properties like friction, surface irregularities, and object momentum. Unlike grasping or lifting tasks, sliding demands precise force calibration and anticipation of the object's response to contact. For evaluation, the agent's learned policy is tested across 50 episodes with different goal locations, assessing its ability to generalize over varied configurations. Training goals are randomly generated from the environment, helping the agent explore diverse sliding trajectories to improve robustness across different scenarios.

\subsection{Block and Pen Rotation} 
In this task, a robotic hand must manipulate either a thin pen or a block to achieve specified target rotations. 
We use "HandManipulatePenRotate-v1" and "HandManipulateBlockRotateXYZ-v1" versions of the gymnasium environments. 
Both tasks feature a state space of 61 dimensions, encompassing the robot's joint states, object states, and goal information. The goal space is 7-dimensional, representing the target pose details.
Each episode will have randomized target rotations goal for all axes of the block and for the x and y axes of the pen.
The pen is more challenging to handle due to its tendency to slip, requiring more precise control compared to the block. 
For evaluation, the latest policy is tested 50 episodes for each task, with each episode having a unique random goal. 
In our framework, training goals are also randomly generated from the environment.

\section{Baselines}
\label{supp:baselines}

We first present our overall training framework for goal-conditioned model-based reinforcement learning (MBRL). 
It is important to note that all baselines utilize this training framework, differing only in the strategy employed for collecting trajectories within the real environment. 
Our training framework is based on the implementation of LEXA paper(\cite{mendonca2021discovering}).

\begin{algorithm}[H]
\begin{algorithmic}[1]
 \State\textbf{Input:} Policy $\pi^G$, $\pi^E$, Environment Goal Distribution $G$, World Model $\hat{M}$, reward function $r^G$, $r^E$
    \State $\mathcal{D} \gets \{ \}$ Initialize buffer. 
    \For{Episode $i=1$ to $N_{\text{train}}$} 
        \State \color{red}$\tau \gets$  Collect trajectories$( \ldots )$\color{black}
       \State $\mathcal{D} \gets \mathcal{D} \cup \tau$
       \State Update world model $\hat{M}$ with $\mathcal{D}$
       \State Update $\pi^G$ in imagination with $\hat{M}$ to maximize $r^G$
       \State Update $\pi^E$ in imagination with $\hat{M}$ to maximize $r^E$
       \EndFor
\end{algorithmic}
\caption{General MBRL Training Framework}
\label{alg:mbrl_training}
\end{algorithm}

\subsection{Go-Explore}
\label{supp:go_explore}

Our baselines utilize the state-of-the-art Go-Explore exploration framework, following the implementation detailed in the PEG paper(\cite{hu2023planning}). 
This approach initially employs a goal-conditioned policy $\pi^G$ to get as close as possible to a specified goal $g$, 
a process referred to as the "Go phase." 
Subsequently, an explorer policy $\pi^E$ is used to further explore the environment starting from the final state of the Go phase, known as the "Explore phase."

The quality of the trajectories generated by the Go-Explore strategy largely depends on the selection of the goal $g$ during the Go phase. 
Therefore, establishing an effective mechanism for selecting the Go phase goals is crucial. 
If the chosen goal $g$ is too simple, the explorer will not sufficiently explore the environment. Conversely, if the goal $g$ is too difficult, the goal-achieving policy $\pi^G$ will fail to approach it effectively. 
Thus, the baselines MEGA-G and PEG-G employ different goal selection strategies to determine $g$, guiding the agent to areas with high exploration potential during the Go phase. 
MEGA-G and PEG-G enhance the agent's exploration efficiency by crafting robust exploration strategies, enabling faster learning of the world model with respect to new dynamic transitions and environmental areas. 
We present the pseudocode for Go-Explore in Algorithm~\ref{alg:go-explore}.

\begin{algorithm}[H]
\begin{algorithmic}[1]
    \Function{GO-EXPLORE($g, \pi^G, \pi^E$)}{}
    \State $s_0 \gets$ env.reset()
    \State $\tau \gets \{s_0\}$
    \For{Step $t=1\ to\ T_{\text{Go}}$}
        \State $s_t \gets$  env.step($\pi^G(s_{t-1}, g)$)
        \State $\tau \gets \tau \cup \{s_t\}$
        \If{agent reach $g$}
        \State break
        \EndIf
    \EndFor
    \State $t_e = t$
    \For{Step $t=t_e\ to\ t_e + T_{\text{Explore}}$}
        \State $s_t \gets$  env.step($\pi^E(s_{t-1})$)
        \State $\tau \gets \tau \cup \{s_t\}$
    \EndFor
    \\\Return $\tau$
\EndFunction
\end{algorithmic}
\caption{Go Explore Framework}
\label{alg:go-explore}
\end{algorithm}

\subsection{GC-Dreamer}
GC-Dreamer is the goal-conditioned version of Dreamer(\cite{hafner2019dream, hafner2019learning, hafner2020mastering}), without incorporating any exploration or goal-directed strategies.
It only uses a goal-conditioned policy to collect trajectories, with goals provided by the training environment.

\begin{algorithm}[H]
\begin{algorithmic}[1]
\Function{Collect trajectories($\ldots$)}{}
\State $g \gets$ Returned by environment 
\State $\tau \gets$ Sample a trajectories by $\pi^G$ using goal $g$
    \\\Return $\tau$
\EndFunction
\end{algorithmic}
\caption{GC-Dreamer Goal Sampling}
\label{alg:gc-dreamer}
\end{algorithm}

\subsection{PEG-G}

PEG uses a world model to simulate exploration trajectories and evaluates the exploration potential($P^E(g)$) to identify areas worth exploring.

\begin{equation}
  P^E(g) = \mathbb{E}_{p_{\pi^G(\cdot \vert \cdot,g)(s_T)}}[V^E(s_T)]
\label{eq:exp_potential_1}
\end{equation}

\begin{equation}
  V^E(s_T) = \mathbb{E}_{\pi^E}[\sum_{t=T+1}^{T+T_E} \gamma^{t-T-1} r^E_{t}] 
  \label{eq:exp_potential_4}
\end{equation}

PEG set goal $g$ for the goal-conditioned policy and generalize it to $K$ trajectories using world model. 
$s_T$ denotes the final state of the goal-conditioned trajectory from the "Go phase" of Go-Explore to reach the $g$.
Since the objective in Equation~\ref{eq:exp_potential_1} is not easily computable, as it relies on the final state distribution induced by the target-conditioned 
policy $\pi^g$, which may rapidly change throughout the training process, it's crucial to use the latest estimates for better exploration. 
PEG achieve this by leveraging the learned world model. PEG utilize the learned exploration value function $V_E(s_k^T)$ (Equation~\ref{eq:exp_potential_2}) from the learned world model to 
estimate the exploration value of the final state for each trajectory, and average these estimates.

\begin{equation}
  \mathbb{E}_{p_{\pi^G(\cdot \vert \cdot,g)(s_T)}}[V^E(s_T)] = \frac{1}{K}\sum_{k}^{K}V^E(s_T^k) \text{\qquad where } s^k_T \sim \hat{p}_{\pi^G(\cdot \vert \cdot,g)(\tau)}
\label{eq:exp_potential_2}
\end{equation}

\begin{equation}
  \hat{p}_{\pi^G(\cdot \vert \cdot,g)(\tau)} = p(s_0)[\prod^{T}_{t=1} \hat{M}(s_t|s_{t-1},a_{t-1})\pi^G(a_{t-1}|s_{t-1},g)]
\label{eq:exp_potential_3}
\end{equation}

The goals sampled for evaluating this exploration potential metric in PEG are drawn from a distribution updated by the MPPI method (\cite{williams2015model,nagabandi2020deep}). 
For more details of PEG MPPI update, please refer to the appendix of their paper(\cite{hu2023planning}).PEG-G not only uses goals obtained by optimizing Equation~\ref{eq:exp_potential_2} to guide exploration sampling but also directly samples trajectories using a goal-conditioned policy with goals provided by the environment. 
The sampling alternate between these two strategies as shown in the pseudocode in Algorithm~\ref{alg:PEG-G}.

\begin{algorithm}[H]
\begin{algorithmic}[1]
\Function{Collect trajectories($\ldots$)}{}
\If{episode $i \% 2 = 0$}
    \State $g \gets$ Optimize Equation~\ref{eq:exp_potential_2} with MPPI
    \State $\tau \gets \textit{GO-EXPLORE}(g, \pi^G, \pi^E)$
\Else
    \State $g \gets$ Returned by environment 
    \State $\tau \gets$ Sample a trajectories by $\pi^G$ using goal $g$
    \EndIf
    \\\Return $\tau$
\EndFunction
\end{algorithmic}
\caption{PEG-G Sampling}
\label{alg:PEG-G}
\end{algorithm}

\subsection{MEGA-G}

MEGA (\cite{pitis2020maximum}) employs kernel density estimates (KDE) to assess state densities and selects goals with low densities from the replay buffer.
For the implementation of MEGA, we adopt the model-based MEGA methodology described in the PEG paper without modifications.
The PEG paper has illustrated that their adaptation of MEGA outperforms the original MEGA implementation. 
This entails integrating MEGA's KDE model and incorporating a goal-conditioned value function into the LEXA framework to filter goals based on reachability. 
Similar to PEG-G, MEGA-G switches between utilizing goals from the environment and employing the MEGA goal selection strategy.

\begin{algorithm}[H]
\begin{algorithmic}[1]
\Function{Collect trajectories($\ldots$)}{}
\If{episode $i \% 2 = 0$}
    \State $g \gets \min_{g \in \mathcal{D}} \widehat{p}(g)$
    \State $\tau \gets \textit{GO-EXPLORE}(g, \pi^G, \pi^E)$
\Else
    \State $g \gets$ Returned by environment 
    \State $\tau \gets$ Sample a trajectories by $\pi^G$ using goal $g$
    \EndIf
    \\\Return $\tau$
\EndFunction
\end{algorithmic}
\caption{MEGA-G Goal Sampling}
\label{alg:MEGA-G}
\end{algorithm}

\subsection{\toolhp{}}

We consider the Time-sample Hindsight Waypoints Sampling Strategy from \textbf{Hindsight Planner} (\cite{lai2020hindsight}) as an alternative to the subgoal selection mechanism in \tool{}. 
\toolhp{} selects subgoals at fixed time intervals along trajectories, providing a simple and effective strategy for defining subgoals.
\toolhp{} can still benefit from the framework of \tool{} in navigating between different subgoals. The pseudocode for this baseline is as follows:

\begin{algorithm}[H]
\caption{\toolhp{} Subgoal Picking Strategy} 
\label{alg: MUN-HP-subgoal-picking}
\begin{algorithmic}[1]
\State \textbf{Function:} Subgoals\_fixed\_interval(...)
\State \textbf{Input:} A batch of episodes $B_{egc}$, number of subgoals $N_{subgoals}$
\State $S_{subgoals} \leftarrow$ pick $N_{subgoals}$ states at fixed time intervals from $B_{egc}$
\State $G_{subgoals} \leftarrow$ $\eta(S_{subgoals})$
\State \textbf{return} $G_{subgoals}$
\end{algorithmic}
\end{algorithm}

\begin{algorithm}[H]
\caption{Trainning Frame for \toolhp{}}
\label{alg: MUN-HP}
\begin{algorithmic}[1]
    \State \textbf{Input:} Policy $\pi^G$, World Model $\hat{M}$, reward function $r^G$, subgoals transfer number $N_{s}$, subgoal time limit $T_{s}$
    \State Initialize buffers $D, D_{DAD}, D_{egc}$
    \For{$i = 1$ to $N_{train}$}
        \If{Should Plan Subgoals}
            \State $B_{egc} \leftarrow$ A batch of episodes from $D_{egc}$
            \State $G_{subgoals} \leftarrow$ Subgoals\_fixed\_interval(...) with Algorithm~\ref{alg: MUN-HP-subgoal-picking}
        \EndIf
        \State Initialize empty trajectory $\tau$
        \For{$s = 1$ to $N_{s}$}
            \State $t_s = 0$
            \State $g_s =$ Sample a subgoal randomly from $G_{subgoals}$
            \While{agent has not reached $g_s$ and $t_s < T_{s}$}
                \State Append one step in real environment with $\pi^G$ using goal $g_s$ to $\tau$
                \State $t_s \leftarrow t_s + 1$
            \EndWhile
        \EndFor
        \State $D_{DAD} \leftarrow D_{DAD} \cup \{\tau\}$
        \State $D_{egc} \leftarrow D_{egc} \cup$ Sample a trajectory with $\pi^G$ using goal from training environment
        \State $D \leftarrow D_{DAD} \cup D_{egc}$
        \State Update $\hat{M}$ with $D$ 
        \State Update $\pi^G$ in imagination with $\hat{M}$ to maximize $r^G$
    \EndFor
\end{algorithmic}
\end{algorithm}

\section{Implementation Details}
\label{supp:implementation_details}

\subsection{Farthest Point Sampling (FPS) Algorithm}

\begin{algorithm}[H]
\caption{Farthest Point Sampling (FPS)}
\label{alg:FPS}
\begin{algorithmic}[1]
\Function{FPS}{points, num\_samples}
    \State sampled\_points $\leftarrow$ [ ]
    \State first\_point $\leftarrow$ random.choice(points)
    \State sampled\_points.append(first\_point)
    \State min\_distances $\leftarrow$ [float('inf')] $\times$ len(points)
    \For{each point $p$ in points}
        \State min\_distances[p] $\leftarrow$ distance(p, first\_point)
    \EndFor
    \For{iteration $i = 1$ to num\_samples-1}
        \State farthest\_point\_index $\leftarrow$ argmax(min\_distances)
        \State farthest\_point $\leftarrow$ points[farthest\_point\_index]
        \State sampled\_points.append(farthest\_point)
        \For{each point $p$ in points}
            \State min\_distances[p] $\leftarrow$ min(min\_distances[p], distance(p, farthest\_point))
        \EndFor
    \EndFor
    \State \Return sampled\_points
\EndFunction
\end{algorithmic}
\end{algorithm}

The pseudocode presented in the Algorithm~\ref{alg:FPS} illustrate the process of Farthest Point Selection (FPS) algorithm. The FPS algorithm begins by initializing an empty list called 'sampled\_points' to store the selected points. The process commences by randomly selecting an initial point from the input point set, denoted as 'points', and adding it to 'sampled\_points'. Subsequently, 'min\_distances' is initialized to keep track of the minimum distance from each point to any of the sampled points, with initial values set to infinity.

The core procedure involves iteratively selecting points until the desired number of samples is reached. At each iteration, the algorithm identifies the point in 'points' with the maximum minimum distance to the previously sampled points and includes it in 'sampled\_points'. Concurrently, 'min\_distances' is updated to reflect the recalculated minimum distance of each point to any of the sampled points.

\subsection{Runtime}

\begin{table}[H]
\centering
\caption{Runtimes per experiment.}
\vspace{1ex}
\label{table:runtime}
\begin{tabular}{lcc}
\toprule
              & Total Runtime (Hours) & Total Steps \\ 
\midrule
3-Block Stacking & 70                    & 2.5e6         \\ 
Walker        & 40                    & 1.5e6       \\ 
Ant Maze      & 36                    & 1e6         \\ 
Block Rotation  & 68                    & 2.5e6       \\ 
Pen Rotation    & 68                    & 2.5e6       \\ 
Fetch Slide    & 52                    & 2e6         \\ 
\bottomrule
\end{tabular}
\end{table}

We conduct each experiment on GPU Nvidia A100 and require about 3GB of GPU memory. See table in Table~\ref{table:runtime} for specific running time of \tool{} for different task.
Most of the runtime is consumed by the neural network updates for the policy and the world model, while the time taken by DAD to filter subgoals is minimal.

\subsection{Hyperparameters}
\label{subsec: hyperparameters}
We use the default hyperparameters of the LEXA backbone MBRL agent (e.g., learning rate, optimizer, network architecture) and keep them consistent across all baselines.
\tool{} primarily requires hyperparameter tuning in the following: 1) the number of candidate subgoals stored $N_{subgoals}$; 2) the number of subgoals used for navigation when sampling in the environment $N_s$; and 3) the total episode length $L$ and the maximum number of timesteps allocated for navigating to a specific subgoal $T_s$.
We show these hyperparameters in Table~\ref{table:mun-hyperparameters}.

\begin{table}[H]
\centering
\caption{Hyperparameters of \tool{}.}
\vspace{1ex}
\label{table:mun-hyperparameters}
\begin{tabular}{lcccc}
\toprule
              & $N_{subgoals}$ & $N_s$ &  $L$  &  $T_s$    \\ 
\midrule
3-Block Stacking & 20             & 2     & 150   & 75        \\ 
Walker        & 10             & 2     & 150   & 75        \\ 
Ant Maze      & 20             & 2     & 500   & 250       \\ 
Block Rotation  & 20             & 2     & 150   & 75        \\ 
Pen Rotation    & 20             & 2     & 150   & 75        \\ 
Fetch Slide    & 20             & 2     & 150   & 75        \\ 
\bottomrule
\end{tabular}
\end{table}

\section{Additional Experiments}

\subsection{More subgoals found by DAD}
\label{subs: subgoalsDAD}

\begin{figure}[h]
  \centering
  \includegraphics[width=0.9\textwidth]{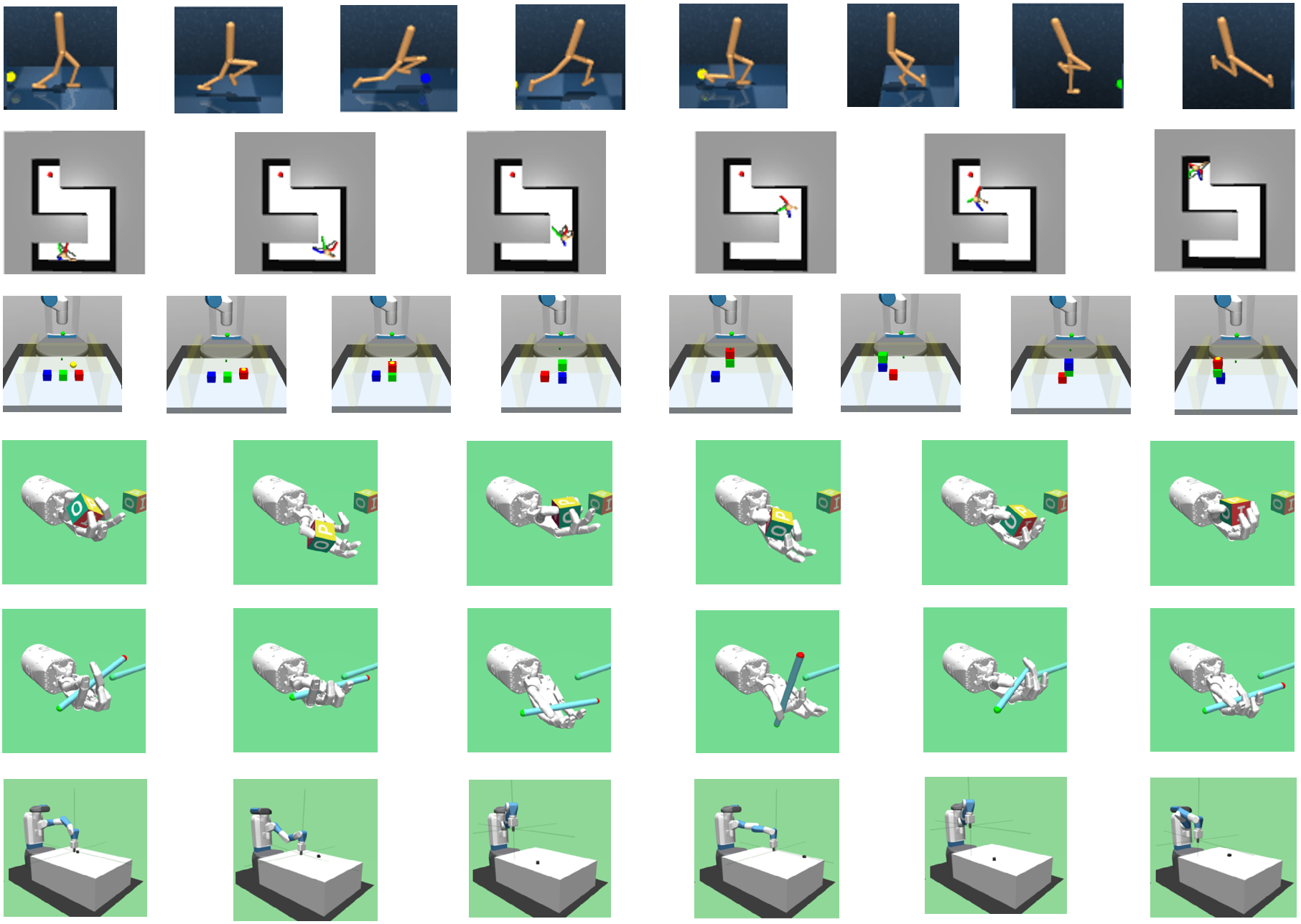}
  \caption{More subgoals found by DAD(Algorithm~\ref{Algorithm2-APS}) in all six environments}
  \label{fig:moresubgoals}
\end{figure}

We visualize several subgoals found by the DAD algorithm during the training process in Fig.~\ref{fig:moresubgoals}. 
In \textbf{Walker}, the first five images show that DAD successfully identifies the crucial joint angles and forces of the Walker robot during its forward locomotion, including standing, striding, jumping, landing, and leg support. 
In the subsequent three images, DAD similarly succeeds in recognizing the key movements of the Walker robot during its backward locomotion. 
%such as reverse ground push-off and rear leg lift, along with their corresponding states.
In \textbf{Ant-Maze}, %despite the complex action and state spaces inherent in quadrupedal ant robots, 
DAD recognizes significant motion variations at corridor corners. %Consequently, it successfully navigates between different rooms by identifying and traveling between such subgoals.
In \textbf{Block Rotation} and \textbf{Pen Rotation}, %the anthropomorphic robotic hand with 24 degrees of freedom need to munipulate the block or the pen to a random target rotation for them. 
%Due to the dexterity of fingers and their intricate joint control, APS encounters some of the most daunting challenges among the six environments. 
DAD is able to identify crucial finger movements subgoals for rotating objects. 
%In instances where certain subgoals offer few assistance in rotating objects, APS gradually diminishes the discovery of such useless subgoals during training as the robotic hand progressively masters object manipulation.
%
In \textbf{3-Block Stacking}, DAD successfully identifies crucial state transitions required during the stacking process.
%of three blocks by leveraging the diversity among actions  and recognizing the set of subgoals corresponding to actions with maximal dissimilarity. 
These critical subgoals include block grasping, lifting, horizontal movement, vertical movement, and gripper release.
%Based on the above example of APS successfully discovering subgoals, we can conclude that APS can efficiently, straightforwardly, and stably identifies the subgoal states required for completing tasks.

\subsection{Navigation Experiments}
\label{subs: more navigation experiments}

We do the extend navigation experiments on 3-Block Stacking, Ant Maze, and Walker environments to see if the \tool{} can learn a better world model to navigate to unconstrained goals from unconstrained start state compared to other baselines.
In the 3-Block Stacking task, we use a set of 15 goals that represent various critical states in the block-stacking process. 
These goals serve as candidates for both initial states and endpoint goals, resulting in a total of 225 unique combinations of initial states and endpoint goals for each evaluation episode. 
For each combination, we conduct 10 repeated evaluations, ultimately computing the average success rate across 2250 evaluation trajectories.
Our goal is to assess whether \tool{} can effectively achieve a random goal when the agent starts from an arbitrary state. 
This evaluation inherently includes both the forward and reverse processes of stacking blocks, determining whether an agent that can stack blocks is also capable of returning the stacked blocks to an intermediate state.
In the Ant Maze environment, we use 32 different positions within the maze as a candidate set for starting and goal positions. Evaluating navigation between these positions allows for a comprehensive assessment of the Ant Robot's world model learning for the maze structure. This evaluation not only measures its ability to reach the final room but also its capability to return to previous rooms from intermediate positions. 
We evaluate 1024 combinations of starting and goal positions, conducting 10 evaluations for each combination, resulting in an average success rate computed over 10,240 experiments.
In the Walker environment, we use all evaluation goals ($\pm13, \pm16, \pm19, \pm22, \pm25, \pm28$) as a candidate set for starting and goal positions. 
This set can form a total of 144 different combinations of starting and goal positions, providing a thorough assessment of the Walker robot's ability to move forward and backward, as well as its precision in position judgment.
See Table~\ref{table:navigation-experiment-3-block},~\ref{table:navigation-experiment-ant},~\ref{table:navigation-experiment-walker} for specific results of \tool{} and all baselines.

% Table for 3-Block Stacking
\begin{table}[H]
\centering
\caption{Success rate of navigation experiments on 3-Block Stacking}
\vspace{1ex}
\label{table:navigation-experiment-3-block}
\begin{tabular}{lcc}
\toprule
             & Environment   & Success rate \\ 
\midrule
\tool{}        & 3-Block Stacking & 95\%         \\ 
\toolhp{}      & 3-Block Stacking & 81\%         \\ 
GC-Dreamer   & 3-Block Stacking & 56\%         \\ 
MEGA-G       & 3-Block Stacking & 42\%         \\ 
PEG-G        & 3-Block Stacking & 47\%         \\ 
\bottomrule
\end{tabular}
\end{table}

% Table for Ant Maze
\begin{table}[H]
\centering
\caption{Success rate of navigation experiments on Ant Maze}
\vspace{1ex}
\label{table:navigation-experiment-ant}
\begin{tabular}{lcc}
\toprule
             & Environment & Success rate \\ 
\midrule
\tool{}        & Ant-Maze    & 96\%         \\ 
\toolhp{}      & Ant-Maze    & 89\%         \\ 
GC-Dreamer   & Ant-Maze    & 75\%         \\ 
MEGA-G       & Ant-Maze    & 94\%         \\ 
PEG-G        & Ant-Maze    & 93\%         \\ 
\bottomrule
\end{tabular}
\end{table}

% Table for Walker
\begin{table}[H]
\centering
\caption{Success rate of navigation experiments on Walker}
\vspace{1ex}
\label{table:navigation-experiment-walker}
\begin{tabular}{lcc}
\toprule
             & Environment & Success rate \\ 
\midrule
\tool{}        & Walker      & 89\%         \\ 
\toolhp{}      & Walker      & 73\%         \\ 
GC-Dreamer   & Walker      & 67\%         \\ 
MEGA-G       & Walker      & 81\%         \\ 
PEG-G        & Walker      & 62\%         \\ 
\bottomrule
\end{tabular}
\end{table}

We observe that \tool{} significantly outperforms other baselines in navigation experiments across all three environments, demonstrating its exceptional contribution to learning comprehensive world models and policies.

\subsection{World Model Assessment}
\label{subs: wm assessment}

Table~\ref{table: One-step Model Prediction Error} shows the single-step prediction error of learned world models. We randomly sample $1e4$ state transition tuples $(s_i, a_i, s_{i+1})$
within the replay buffers from all of our baselines (MUN, MUN-noDAD, GC-Dreamer, MEGA-G, and PEG-G) to form a validation dataset. 
Table~\ref{table: One-step Model Prediction Error} reports the mean squared error on this dataset.

Table~\ref{table: Compound Model Prediction Error} shows the compounding error (multistep prediction error) of learned world models for evaluation when generating the same length simulated trajectories. 
More specifically, assume a real trajectory of length $h$ is denoted as $(s_0, a_0, s_1, ..., s_h)$. For a learned model $M$, we sample from $s_0$ and generate forward rollouts $(\hat{s}_0, a_0, \hat{s}_1, ..., \hat{s}_h)$
where $\hat{s}_0 = s_0$ and for $i \leq 0$, $\hat{s}_{i+1} = M(\hat{s}_0, a_i)$. Then the corresponding compounding error of $M$ is defined as $\frac{1}{h} \sum_{i=1}^{h} \left\| \hat{s}_i - s_i \right\|_2^2$. We set $h$ to be the maximum number of timesteps in our environments. We evaluated the compounding prediction error of the learned world models by generating 500 trajectories for each benchmark, simulated on both the models and the real environments.

\begin{figure}[h] 
  \centering
    \subfigure[Trajectories 3-Block Stacking]{\includegraphics[width=0.4\textwidth]{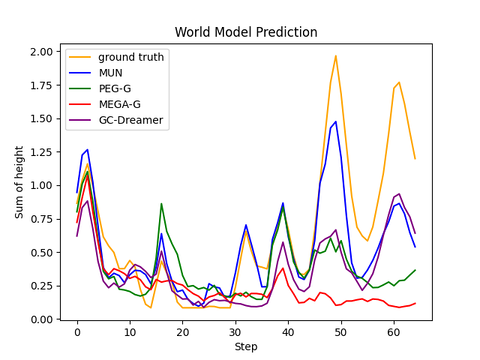}}
    \subfigure[Trajectories Block Rotation]{\includegraphics[width=0.4\textwidth]{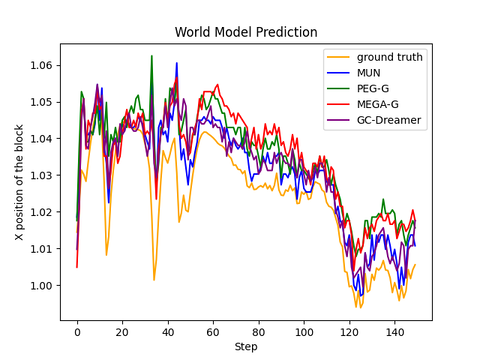}}
  \caption{Fig(a) and Fig(b) illustrate the imagined and real environment trajectories for 3-Block Stacking and Block Rotation respectively, starting from the same initial state. Among the baselines, MUN demonstrates the smallest compound model error with respect to the ground truth trajectories. The X-axis represents the trajectory steps. In Fig(a), the Y-axis represents the sum of the heights of the three blocks. MUN's world model outperforms other methods in predicting the correct locations of the three blocks. In Fig(b), the Y-axis represents the position of the block in the x coordinate. MUN's world model outperforms other methods in predicting the correct position of the block.}
  \vspace{-10pt}
  \label{fig: compound error episode}
\end{figure}

In Tables ~\ref{table: One-step Model Prediction Error} and ~\ref{table: Compound Model Prediction Error}, we used the final world models trained by all methods after the same number of environment interaction steps. These results provide a quantitative comparison of the world model prediction quality between MUN and the baselines across our benchmarks. The world models trained by MUN show a much smaller generalization gap to the real environment compared to goal-conditioned Dreamer (and the other baselines). Consequently, MUN can effectively leverage these world models to train control policies that generalize well to the real environment. This explains the superior task success rates of MUN compared to the baselines in our experiment. Fig~\ref{fig: compound error episode} also provides more information about the world model compound prediction error.

\begin{table}[H]
\centering
\caption{One-step Model Prediction Error.}
\vspace{1ex}
\label{table: One-step Model Prediction Error}
\begin{tabular}{lccccc}
\toprule
              & MUN   & MUN-noDAD & PEG-G & MEGA-G & GC-Dreamer \\ 
\midrule
Ant Maze      & 1.6740 & 1.9751    & 2.1154 & 2.2416 & 2.9666    \\ 
Walker        & 0.8165 & 0.9971    & 1.4759 & 1.2353 & 2.1824    \\ 
3-Block Stacking & 0.0070 & 0.0071    & 0.0476 & 0.0853 & 0.0392    \\ 
Rotate Block  & 1.0570 & 1.5609    & 1.7753 & 1.9433 & 2.3723    \\ 
Rotate Pen    & 0.6708 & 1.1999    & 1.9622 & 2.8598 & 1.8359    \\ 
Fetch Slide    & 0.0094 & 0.0108    & 0.0132 & 0.0164 & 0.0169    \\ 
\bottomrule
\end{tabular}
\end{table}

\begin{table}[H]
\centering
\caption{Compound Model Prediction Error.}
\vspace{1ex}
\label{table: Compound Model Prediction Error}
\begin{tabular}{lccccc}
\toprule
              & MUN   & MUN-noDAD & PEG-G & MEGA-G & GC-Dreamer \\ 
\midrule
Ant Maze      & 18.83 & 22.42     & 29.42 & 23.69  & 40.36      \\ 
Walker        & 13.03 & 16.72     & 26.54 & 21.21  & 39.72      \\ 
3-Block Stacking & 0.45  & 0.55      & 0.70  & 0.95   & 0.94       \\ 
Rotate Block  & 11.55 & 12.86     & 14.38 & 14.13  & 15.06      \\ 
Rotate Pen    & 4.63  & 6.10      & 7.40  & 9.85   & 9.36       \\ 
Fetch Slide    & 1.687  & 1.648      & 2.195  & 2.856   & 2.304       \\ 
\bottomrule
\end{tabular}
\end{table}

%%%%%%%%%%%%%%%%%%%%%%%%%%%%%%%%%%%%%%%%%%%%%%%%%%%%%%%%%%%%

\newpage
\section*{NeurIPS Paper Checklist}

\begin{enumerate}

\item {\bf Claims}
    \item[] Question: Do the main claims made in the abstract and introduction accurately reflect the paper's contributions and scope?
    \item[] Answer: \answerYes{} % Replace by \answerYes{}, \answerNo{}, or \answerNA{}.
    \item[] Justification: Our abstract and introduction accurately represent the primary contributions of the paper, which include the development of the \tool{} algorithm designed to improve goal-conditioned reinforcement learning through enhanced world modeling and exploration capabilities. The introduction outlines the key challenges in GCRL, specifically with sparse rewards, and how \tool{} addresses these by facilitating effective state transitions between arbitrary subgoal states in the replay buffer. These claims are well-supported by the theoretical underpinnings and experimental results presented in the paper, reflecting the scope and impact of the proposed method. 

\item {\bf Limitations}
    \item[] Question: Does the paper discuss the limitations of the work performed by the authors?
    \item[] Answer: \answerYes{} % Replace by \answerYes{}, \answerNo{}, or \answerNA{}.
    \item[] Justification: Our paper thoroughly discusses the limitations of the \tool{} framework in Appendix. We highlight the dependency on an efficient strategy for discovering key states, pointing out that while DAD is effective, it will also identify irrelevant states in tasks with complex action spaces or weak correlations between goal space and action space. Additionally, we mention the potential for applying \tool{} to model-free reinforcement learning methods, which do not require learning a world model and have simpler architectures. 

\item {\bf Theory Assumptions and Proofs}
    \item[] Question: For each theoretical result, does the paper provide the full set of assumptions and a complete (and correct) proof?
    \item[] Answer: \answerNA{} % Replace by \answerYes{}, \answerNo{}, or \answerNA{}.
    \item[] Justification: Our paper does not include theoretical results.
    \item[] Guidelines: In the Experiment section and the Appendix of our paper, we provide a detailed description of our experimental procedures and configurations. 
    This includes all sources and modifications of the test environments, pseudocode and implementation methods for all baselines, the equipment and memory used, 
    as well as the specific values of the required hyperparameters. Additionally, we have open-sourced our code, which can be found in the Reproducibility Statement section.
    \begin{itemize}
        \item The answer NA means that the paper does not include theoretical results. 
        \item All the theorems, formulas, and proofs in the paper should be numbered and cross-referenced.
        \item All assumptions should be clearly stated or referenced in the statement of any theorems.
        \item The proofs can either appear in the main paper or the supplemental material, but if they appear in the supplemental material, the authors are encouraged to provide a short proof sketch to provide intuition. 
        \item Inversely, any informal proof provided in the core of the paper should be complemented by formal proofs provided in appendix or supplemental material.
        \item Theorems and Lemmas that the proof relies upon should be properly referenced. 
    \end{itemize}

    \item {\bf Experimental Result Reproducibility}
    \item[] Question: Does the paper fully disclose all the information needed to reproduce the main experimental results of the paper to the extent that it affects the main claims and/or conclusions of the paper (regardless of whether the code and data are provided or not)?
    \item[] Answer: \answerYes{} % Replace by \answerYes{}, \answerNo{}, or \answerNA{}.
    \item[] Justification: In the Experiment section and appendix of our paper, we elaborate on the procedure and configuration of our experiments. 
    This includes the sources and modifications of all testing environments, pseudo code and implementation methods for all baselines, the devices and memory utilized, as well as specific values of hyperparameters employed. 
    Concurrently, we have open-sourced our code; please refer to the Reproducibility Statement section for further details.

\item {\bf Open access to data and code}
    \item[] Question: Does the paper provide open access to the data and code, with sufficient instructions to faithfully reproduce the main experimental results, as described in supplemental material?
    \item[] Answer: \answerYes{} % Replace by \answerYes{}, \answerNo{}, or \answerNA{}.
    \item[] Justification: As we mentioned in the previous justification, we have not only open-sourced our code but also provided detailed steps and settings for reproducing our main experimental results. 
    In the Experiment section and Appendix, we elaborate on the sources and modifications of the environments, baseline implementation details, and \tool{} implementation specifics.


\item {\bf Experimental Setting/Details}
    \item[] Question: Does the paper specify all the training and test details (e.g., data splits, hyperparameters, how they were chosen, type of optimizer, etc.) necessary to understand the results?
    \item[] Answer: \answerYes{} % Replace by \answerYes{}, \answerNo{}, or \answerNA{}.
    \item[] Justification: We provide comprehensive details regarding the hyperparameters essential for understanding the experiments, including those specific to our \tool{} framework. 
    The table presented (Fig~\ref{table:mun-hyperparameters}) outlines these hyperparameters for each task, facilitating reproducibility and comparison.

\item {\bf Experiment Statistical Significance}
    \item[] Question: Does the paper report error bars suitably and correctly defined or other appropriate information about the statistical significance of the experiments?
    \item[] Answer: \answerYes{} % Replace by \answerYes{}, \answerNo{}, or \answerNA{}.
    \item[] Justification: We conducted each experiment a minimum of five times using different random seeds, and upon plotting the results, as demonstrated in the Experiment section, we incorporated the experimental error. 
    The solid line denotes the average success rate, while the shaded region signifies the standard deviation among the repeated experimental outcomes.

\item {\bf Experiments Compute Resources}
    \item[] Question: For each experiment, does the paper provide sufficient information on the computer resources (type of compute workers, memory, time of execution) needed to reproduce the experiments?
    \item[] Answer: \answerYes{} % Replace by \answerYes{}, \answerNo{}, or \answerNA{}.
    \item[] Justification: We clearly specifies the computer resources (Nvidia A100 GPU) and the amount of GPU memory required (approximately 3GB). 
    Additionally, we provides detailed information on the runtime of each experiment in Appendix.
    
\item {\bf Code Of Ethics}
    \item[] Question: Does the research conducted in the paper conform, in every respect, with the NeurIPS Code of Ethics \url{https://neurips.cc/public/EthicsGuidelines}?
    \item[] Answer: \answerYes{} % Replace by \answerYes{}, \answerNo{}, or \answerNA{}.
    \item[] Justification: The research conducted in our paper aligns with the NeurIPS Code of Ethics. We have thoroughly reviewed the guidelines and ensured that our research adheres to ethical standards. Additionally, we have implemented measures to safeguard anonymity and comply with pertinent laws and regulations.

\item {\bf Broader Impacts}
    \item[] Question: Does the paper discuss both potential positive societal impacts and negative societal impacts of the work performed?
    \item[] Answer: \answerNA{} % Replace by \answerYes{}, \answerNo{}, or \answerNA{}.
    \item[] Justification: Our research aims to address the exploration problem in Reinforcement Learning (RL) within the GCRL environment. It is currently in the theoretical research stage and has minimal societal impact.
    
\item {\bf Safeguards}
    \item[] Question: Does the paper describe safeguards that have been put in place for responsible release of data or models that have a high risk for misuse (e.g., pretrained language models, image generators, or scraped datasets)?
    \item[] Answer: \answerNA{} % Replace by \answerYes{}, \answerNo{}, or \answerNA{}.
    \item[] Justification: Our paper poses no such risks. 

\item {\bf Licenses for existing assets}
    \item[] Question: Are the creators or original owners of assets (e.g., code, data, models), used in the paper, properly credited and are the license and terms of use explicitly mentioned and properly respected?
    \item[] Answer: \answerYes{} % Replace by \answerYes{}, \answerNo{}, or \answerNA{}.
    \item[] Justification: Our paper properly credits the creators or original owners of assets used, including code, data, and models. 
    The licenses and terms of use are explicitly respected. Specifically, we cite the original papers for code packages or datasets used, state the version of the assets, and include URLs where possible.

\item {\bf New Assets}
    \item[] Question: Are new assets introduced in the paper well documented and is the documentation provided alongside the assets?
    \item[] Answer: \answerYes{} % Replace by \answerYes{}, \answerNo{}, or \answerNA{}.
    \item[] Justification: We have documented our code and provided detailed instructions on its usage, licenses, and permissible scope of use. 
    Additionally, we have included the documentation alongside the assets to ensure accessibility and clarity for users.

\item {\bf Crowdsourcing and Research with Human Subjects}
    \item[] Question: For crowdsourcing experiments and research with human subjects, does the paper include the full text of instructions given to participants and screenshots, if applicable, as well as details about compensation (if any)? 
    \item[] Answer: \answerNA{} % Replace by \answerYes{}, \answerNo{}, or \answerNA{}.
    \item[] Justification: Our paper not involve crowdsourcing nor research with human subjects.

\item {\bf Institutional Review Board (IRB) Approvals or Equivalent for Research with Human Subjects}
    \item[] Question: Does the paper describe potential risks incurred by study participants, whether such risks were disclosed to the subjects, and whether Institutional Review Board (IRB) approvals (or an equivalent approval/review based on the requirements of your country or institution) were obtained?
    \item[] Answer: \answerNA{} % Replace by \answerYes{}, \answerNo{}, or \answerNA{}.
    \item[] Justification: Our paper does not involve crowdsourcing nor research with human subjects.

\end{enumerate}
\end{document}